\documentclass[10pt,twocolumn,letterpaper]{article}

\usepackage[pagenumbers]{cvpr} % To force page numbers, e.g. for an arXiv version

\usepackage{amsmath,amssymb}
\usepackage{graphicx}   
\usepackage{pifont}
\usepackage{makecell}
\usepackage{algorithm2e}
\usepackage{multirow}

\definecolor{cvprblue}{rgb}{0.21,0.49,0.74}
\usepackage[pagebackref,breaklinks,colorlinks,allcolors=cvprblue]{hyperref}

\title{Mesh-Pro: Asynchronous Advantage-guided Ranking Preference Optimization for Artist-style Quadrilateral Mesh Generation}

\author{
Zhen Zhou\textsuperscript{1,2, *}, Jian Liu\textsuperscript{2,3, *}, Biwen Lei\textsuperscript{2}, Jing Xu\textsuperscript{2}, Haohan Weng\textsuperscript{2}, Yiling Zhu\textsuperscript{2}, Zhuo Chen\textsuperscript{2},\\
Junfeng Fan\textsuperscript{1}, Yunkai Ma\textsuperscript{1,$\dagger$}, Dazhao Du\textsuperscript{3}, Song Guo\textsuperscript{3,$\dagger$},Fengshui Jing\textsuperscript{1},Chunchao Guo\textsuperscript{2}\\
\textsuperscript{1}Institute of Automation, Chinese Academy of Sciences, \textsuperscript{2}Tencent Hunyuan,\\ 
\textsuperscript{3}Hong Kong University of Science and Technology\\
{\tt\small zhouzhen2021@ia.ac.cn, \textsuperscript{*}Equal contribution, \textsuperscript{$\dagger$}Corresponding author.}
}

\usepackage{fancyhdr}

\fancypagestyle{arxivheader}{
    \fancyhf{} % 清空默认的页眉页脚
    
    \lhead{
        \raisebox{-0.2\height}{\includegraphics[height=12pt]{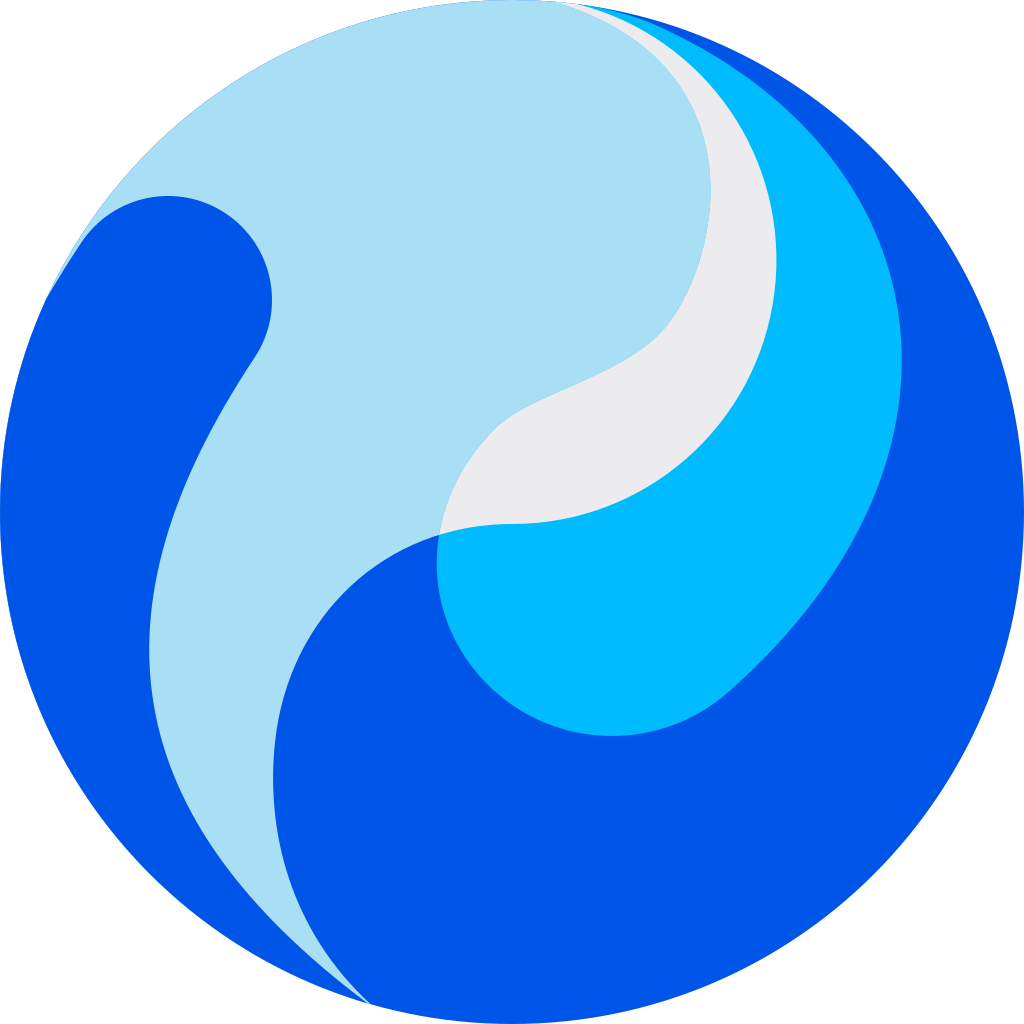}}
        \hspace{3pt} % 建议留一点间距，比 0pt 视觉效果更好
        \textsf{Tencent Hunyuan} 
    } 
    
    \rhead{
        \textsf{Published as a conference paper at CVPR 2026}
    }
    
}
\begin{document}

\twocolumn[{
\maketitle
\thispagestyle{arxivheader}
\begin{center}
\includegraphics[width=0.9\linewidth]{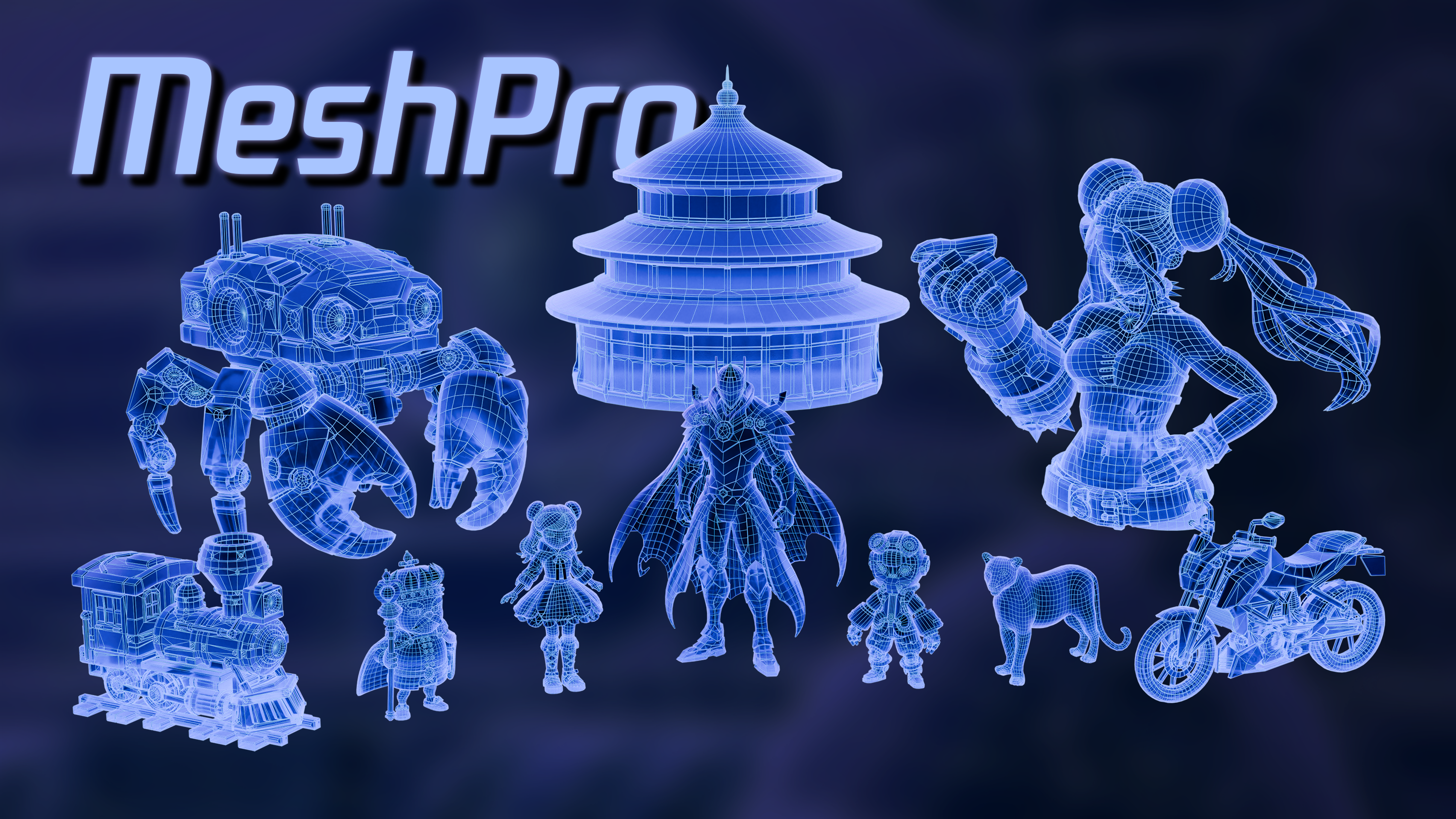}
\end{center}
\vspace{-0.5cm}
\captionsetup{type=figure}
\captionof{figure}{%
\textbf{Mesh-Pro} generates artist-style quadrilateral-dominated meshes with diversity, high fidelity, and topological quality.
}
\label{fig:teaser}
\vspace{0.3cm}
}]

% 在主干论文内容开始前，暂停将章节写入目录
\addtocontents{toc}{\protect\setcounter{tocdepth}{0}}

\begin{abstract}
Reinforcement learning (RL) has demonstrated remarkable success in text and image generation, yet its potential in 3D generation remains largely unexplored. Existing attempts typically rely on offline direct preference optimization (DPO) method, which suffers from low training efficiency and limited generalization. In this work, we aim to enhance both the training efficiency and generation quality of RL in 3D mesh generation. Specifically, (1) we design the first asynchronous online RL framework tailored for 3D mesh generation post-training efficiency improvement, which is 3.75$\times$ faster than synchronous RL. (2) We propose Advantage-guided Ranking Preference Optimization (ARPO), a novel RL algorithm that achieves a better trade-off between training efficiency and generalization than current RL algorithms designed for 3D mesh generation, such as DPO and group relative policy optimization (GRPO). (3) Based on asynchronous ARPO, we propose Mesh-Pro, which additionally introduces a novel diagonal-aware mixed triangular-quadrilateral tokenization for mesh representation and a ray-based reward for geometric integrity. Mesh-Pro achieves state-of-the-art performance on artistic and dense meshes. 
\end{abstract}
\vspace{-0.4cm}
\section{Introduction}
High-quality 3D mesh generation is vital for applications like gaming and embodied intelligence \cite{MeshFormer, PolyGen, MeshGPT, LIRA}, as mesh quality directly dictates visual fidelity and realism. To generate higher-quality meshes, many native mesh generation methods \cite{PivotMesh, chen2024meshanythingv2, xu2025meshmosaic} have emerged. Among these, approaches based on geometry sequence tokenizers \cite{chen2025meshxl, chen2024meshanything, tang2025edgerunner, lionar2025treemeshgpt, weng2025scaling}, such as Meshtron \cite{hao2024meshtron}, have demonstrated powerful modeling capabilities. However, training through supervised learning often results in undesirable artifacts such as holes, non-manifold surfaces, tiny facets, and disorganized topology. Recently, reinforcement learning (RL) has been employed to improve 3D mesh generation \cite{zhao2025deepmesh, liu2025meshrft, liu2025quadgpt}. This is accomplished by further training the pretrained network under the guidance of human-constructed preference pairs or rule-based reward signals. Specifically, offline Direct Preference Optimization (DPO) \cite{DPO} is used to improve topology quality and bring the mesh quality closer to the standard of professional artists.

% The training efficiency and performance of offline RL are limited, making it necessary to extend to online RL training. However, in autoregressive 3D mesh generation frameworks, the training cost of synchronous online RL is extremely high. This is mainly because the rollout data have highly varying token lengths, so synchronous training suffers from long waiting times between policy updates and severe GPU idle periods. Many asynchronous RL frameworks have been developed for autoregressive text modeling, such as AREAL \cite{AREAL}, VeRL \cite{HybridFlow} and OpenRLHF \cite{OpenRLHF}, but they are highly specialized and not suitable for autoregressive 3D mesh generation. Furthermore, existing works construct pairwise data samples and apply DPO for RL fine-tuning. DPO performs implicit reward modeling, featuring a smoother and more convex objective that leads to faster and stable model convergence. Since DPO cannot explicitly capture complex reward function distributions, it tends to exhibit poor generalization to out-of-distribution data. 

Offline DPO in mesh generation post-training has two main problems: (1) Offline training relies on pre-constructed static data, making it unable to dynamically update the policy and data, which leads to low training efficiency. Therefore, it is necessary to explore online RL. In autoregressive 3D mesh generation frameworks, the training cost of synchronous online RL is extremely high. This is mainly because the rollout data have highly varying token lengths, so synchronous training suffers from long waiting times between policy updates and severe GPU idle periods. Many asynchronous RL frameworks have been developed for autoregressive text modeling, such as AREAL \cite{AREAL}, VeRL \cite{HybridFlow} and OpenRLHF \cite{OpenRLHF}, but they are highly specialized and not suitable for autoregressive 3D mesh generation. (2) DPO performs implicit reward modeling, featuring a smoother and more convex objective that leads to faster and stable model convergence. Since DPO cannot explicitly capture complex reward function distributions, it tends to exhibit poor generalization to out-of-distribution data. 

In this paper, we design the first asynchronous online RL framework that is better suited for 3D mesh generation. The proposed asynchronous online RL framework improves training efficiency by more than \textbf{3.75}$\times$ over synchronous RL in large-scale distributed settings. Moreover, to improve the generalization performance of RL in mesh generation, we attempt to explicitly learn the reward distribution using methods such as Group Relative Policy Optimization (GRPO) \cite{GRPO}. Our empirical results find that, constrained by the capabilities of the pretrained foundation model and the complexity of the reward distribution, GRPO exhibits low exploration-exploitation efficiency and slow convergence, which limits the mesh quality. Hence, we further propose \textbf{Advantage-guided Ranking Preference Optimization (ARPO)}. ARPO utilizes ranking preference optimization to achieve fast and stable training convergence, while explicitly introducing advantage function guidance to enhance generalization ability, achieving a better trade-off between training efficiency and generalization.

Based on asynchronous ARPO, we propose \textbf{Mesh-Pro}. To achieve an artist-like topology, Mesh-Pro utilizes an end-to-end autoregressive framework to directly generate native quadrilateral meshes \cite{liu2025quadgpt}. For mixed triangle-quad mesh tokenization, prior methods \cite{liu2025quadgpt} declare the face type (triangle or quad) with a leading special token and preclude a truly consistent canonical ordering, resulting in geometric artifacts and structural defects. To end this, Mesh-Pro introduces a novel diagonal-aware mesh tokenization scheme for mixed triangle-quad meshes. Moreover, we also introduce a ray-based reward mechanism for asynchronous ARPO to reduce broken ratio of mesh generation. In summary, our major contributions are as follows:

\begin{enumerate}
\item We design the first asynchronous online RL framework for mesh generation post-training efficiency improvement, achieving 3.75$\times$ faster than synchronous RL.
\item We propose Advantage-guided Ranking Preference Optimization (ARPO), a novel RL algorithm that achieves a better trade-off between training efficiency and generalization, and generates higher-quality quad meshes.
\item We propose Mesh-Pro, which integrates asynchronous ARPO, a novel diagonal-aware mixed triangular-quadrilateral tokenization method, and a ray-based reward for geometric integrity, outperforming existing methods in artist-style mesh generation.
\end{enumerate}    
\section{Related Work}
\label{sec:related_work}

\begin{figure*}[t]
    \centering
    \includegraphics[width=1.0\textwidth]{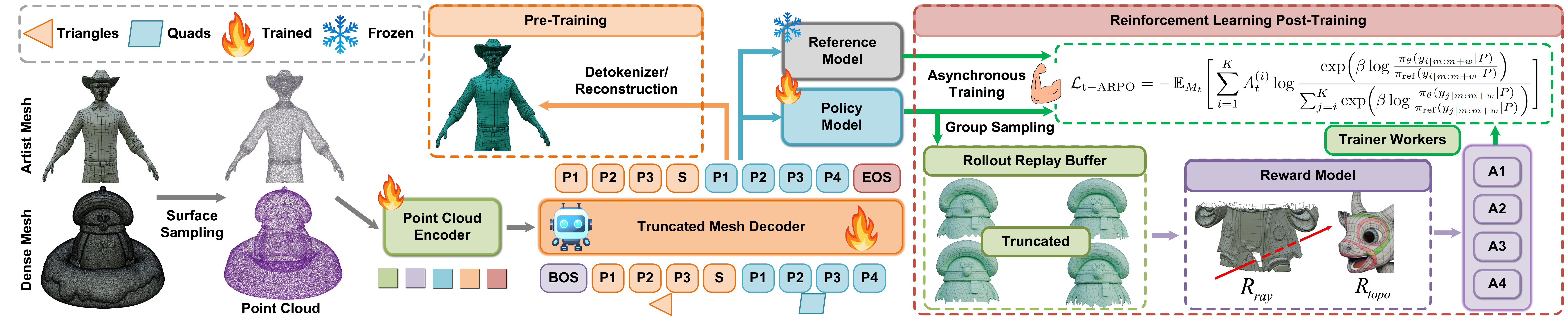}
    \caption{
    \textbf{Architecture Overview.} Mesh-Pro begins by sampling point clouds from the input dense and artist meshes. The features from the point cloud encoder are then passed to an auto-regressive Hourglass Transformer \cite{hao2024meshtron} for mesh decoding. This decoder is trained with truncation to output triangle-quad tokens. The pre-training objective is to reconstruct the input mesh. Subsequently, asynchronous ARPO is used for RL post-training to generate high-quality, well-structured meshes, guided by ray and topological rewards.
    }
    \label{fig:overview}
\end{figure*}

\subsection{3D Mesh Generation}

The 3D mesh generation landscape has been largely shaped by methods centered on continuous neural representations like vecsets~\citep{zhang20233dshape2vecset, zhang2024clay, li2024craftsman, zhao2023michelangelo, wu2024direct3d, li2025triposg, zhao2025hunyuan3d, lai2025unleashing, chen2025ultra3d} or sparse voxels~\citep{xiang2024structured, ye2025hi3dgen, wu2025direct3d-s2, he2025sparseflex, li2025sparc3d, li2025voxhammer, ye2025nano3d}. These approaches are universally constrained by a reliance on iso-surfacing algorithms~\cite{lorensen1998marching}, which inherently produce dense, topologically unstructured triangular meshes. While some techniques accelerate field prediction to generate quadrilaterals~\citep{DL2quadMesh2021, li2025point2quad, NeurCross2024, dong2025crossgen, liu2025neuframeq}, they are not end-to-end generative frameworks and their outputs often lack the intentional edge flow prized by artists. In response, autoregressive models have emerged as a powerful paradigm~\cite{PolyGen, chen2025meshxl, chen2024meshanything, tang2025edgerunner, hao2024meshtron, weng2025scaling, lionar2025treemeshgpt, liu2025freemesh, song2025mesh, kim2025fastmesh, xu2025meshmosaic, chen2025xspecmesh, zhao2025deepmesh, liu2025meshrft, liu2025quadgpt}, operating directly on raw mesh data. Key advancements, such as Meshtron's\cite{hao2024meshtron} scalable transformer architecture and QuadGPT's\cite{liu2025quadgpt} native quad representation with RL-based alignment, have significantly pushed the SOTA. Nevertheless, insufficient quadrilateral mesh tokenization and simplistic RL post-training limit mesh generation efficiency and topology quality.

\subsection{Reinforcement Learning in Mesh Generation}
Inspired by the significant performance improvements of RL in text and image generation \cite{NCA, KTO, Flow-grpo, ImageReward, RL_1, RL_2}, several studies \cite{zhao2025deepmesh, liu2025meshrft, liu2025quadgpt} have also explored its application in the 3D mesh generation domain. DeepMesh \cite{zhao2025deepmesh} used artificially constructed preference data to support the post-training of DPO. Mesh-RFT \cite{liu2025meshrft} and QuadGPT \cite{liu2025quadgpt} introduced a rule-based reward mechanism to improve the geometric integrity and topology quality. However, they use offline DPO, resulting in low training efficiency and insufficient generalization ability. To address this, we propose asynchronous ARPO to enhance RL training efficiency and generalization performance, ultimately generating well-structured artist-style quadrilateral meshes.  
\section{Mesh-Pro}
We introduce Mesh-Pro, starting with the pre-trained model in \cref{sec:pretrain}, followed by the asynchronous RL framework in \cref{sec:async_RL}, ARPO in \cref{sec:ARPO}, and finally reward design in \cref{sec:reward}. The architecture is shown in \cref{fig:overview}.

\subsection{Mesh Generation Pre-Training}
\label{sec:pretrain}

% 这里解释一下这种设计带来的好处，数据集中三四边面混合。 采用旧序列化的第一个顶点就得决定生成三/四边面，一旦选择四边面，就不能反悔，预测压力大，容易破损。而新序列化可以先生成一个三角面，有前面三个顶点的信息，再决定是否生成四边面，就算生成四边面压力也会小一些。 同时我们保证了每次生成面的第一个最低顶点在最前面，sequence 更加 结构化，更容易学习。 虽然引入了扩大了额外的vocab size, 来区分对角线，貌似更难学习，但是实验发现这种级别的词表学习难度并不大。 但实验发现，模型会倾向于偷懒，生成较多special token, 不过在后面的强化中，我们发现这种是可以很好被解决的。

\begin{figure}[t]
    \centering
    \includegraphics[width=0.48\textwidth]{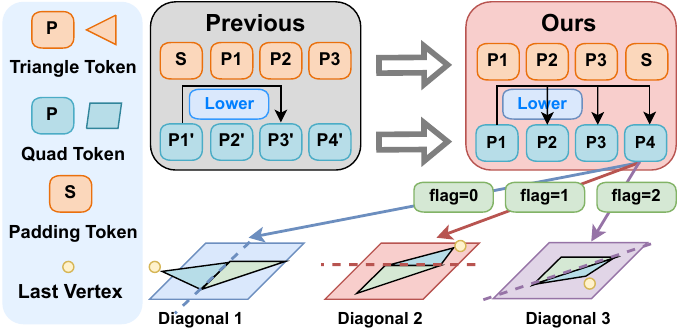}
    \caption{
    Diagonal-Aware Mesh Tokenization. ``P'' denotes vertex tokens. The minimum vertex always appears first in each face (i.e., lower coordinates). Triangles use padding tokens (``S'') at the end, while quads encode diagonal information in the fourth vertex via offset $flag \times 2^{n_{bits}}$ ($flag \in \{0,1,2\}$ for Diagonals {1, 2, 3}). The three edges of the first triangle (in green) can all potentially serve as diagonals. This defers the triangle-vs-quad decision to the last position, reducing prediction pressure. 
    }
    \label{fig:tokenization}
\vspace{-0.2cm}
\end{figure}

% \subsubsection{Diagonal-Aware Mesh Tokenization}
\noindent\textbf{Diagonal-Aware Mesh Tokenization.}
We develop a novel diagonal-aware tokenization scheme for mixed triangle-quad meshes. First, we establish a canonical representation by normalizing vertex coordinates to $[-0.95, 0.95]^3$, applying $n$-bit quantization ($n=10$), and sorting them lexicographically,
\begin{equation}
\tilde{\mathbf{v}}_i = \left\lfloor \frac{(\mathbf{v}_i - \mathbf{v}_{\min})}{(\mathbf{v}_{\max} - \mathbf{v}_{\min})} \cdot 2^n \right\rceil \in [0, 2^n-1]^3.
\end{equation}
The model first generates a three-vertex base triangle, then decides to either terminate with a padding token (creating a triangle) or append a fourth vertex. For quadrilaterals, the internal diagonal is explicitly encoded via an offset, $flag \times 2^{n_{bits}}$, added to the fourth vertex's index, where $flag \in \{0, 1, 2\}$ specifies the diagonal's orientation (see \cref{fig:tokenization}). This ``generate-then-decide'' strategy reduces the predictive burden on the model. This tokenization design addresses critical drawbacks in prior methods (e.g., QuadGPT~\cite{liu2025quadgpt}), which use a leading token that forces a premature commitment to the face type and rely on an inconsistent, non-canonical ordering (only the first vertex is lower than the third vertex). Our approach resolves this by deferring the decision and enforcing a global order where every face sequence begins with its absolute minimum-index vertex. Our tokenization achieves more stable mesh generation. While supervised learning alone can create a bias towards simpler triangular faces, this is effectively rectified in our subsequent asynchronous ARPO phase, which rewards the formation of well-structured quadrilaterals. The full algorithm is detailed in the supplementary materials.

\noindent\textbf{Model Architecture and Training.}
We adopt the Hourglass Transformer~\citep{hao2024meshtron, nawrot2021hierarchical, liu2025quadgpt} that progressively compresses sequences by factors of 4 and 3, capturing global patterns in bottleneck layers while preserving local details. Mesh-Pro is pre-trained autoregressively on
\begin{equation}
\mathcal{L}_{\text{pretrain}} = -\sum_{t=1}^{L} \log p_\theta(\mathbf{s}_t | \mathbf{s}_{<t}),
\end{equation}
where $\mathbf{S} = (\mathbf{s}_1, \ldots, \mathbf{s}_L)$ is the tokenized sequence.

%%%%%%%%%%%%%%%%%%%%%%%%%%%%%%%%%%%%%%%%%%%%%%%%%%%%%%%%%%%%%%%%%%%%%%%%%%%%
\begin{figure}[t]
    \centering
    \includegraphics[width=0.48\textwidth]{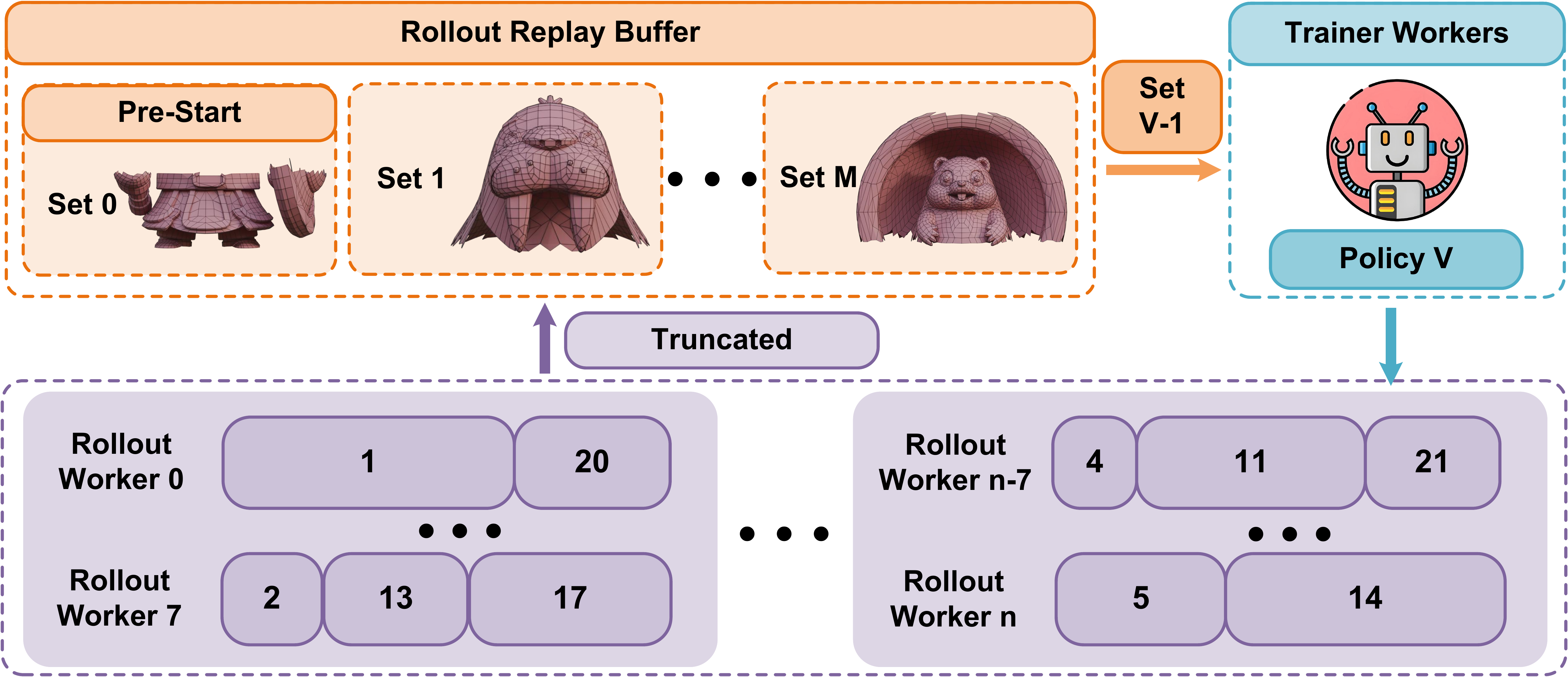}
    \caption{ 
    Asynchronous Online RL Framework.
    }
    \label{fig:Async_RL}
\end{figure}

\subsection{Asynchronous Online RL Framework}
\label{sec:async_RL}
To improve the efficiency of RL post-training for 3D mesh generation, we design an asynchronous online RL framework, as described in Fig.~\ref{fig:Async_RL}. Rollout workers continuously generate training data, while the trainer workers constantly sample data from the rollout replay buffer for policy updates. Every fixed number of training steps, the updated model weights are saved. Meanwhile, the rollout workers periodically fetch the latest weights to ensure that the policies used for generating rollouts do not deviate too far from the current trainer policy. To guarantee that the trainer workers always receive data generated by the most recent policy, outdated rollout data are promptly discarded from the rollout replay buffer. Unlike AREAL~\cite{AREAL}, our asynchronous framework updates model weights without interrupting the rollout process, ensuring that each rollout is generated under a consistent policy. Asynchronous training eliminates significant idle waiting time between rollout and trainer workers, while dynamically updating both the policy and data to improve RL efficiency.

\noindent\textbf{Pre-Start Stage for Stable Training.} To enhance the stability of asynchronous RL training, saving the first policy is crucial. We expect the model to have preliminarily adapted to the reward distribution by the time this first policy is saved, such that the meshes generated by this policy outperform those from the pretrained model policy. Specifically, the rollout workers first collect $S_1$ samples into the rollout replay buffer, after which the trainer workers begin training. The first policy checkpoint is saved after $N_1$ training steps. This process can be regarded as a pre-start stage within the asynchronous RL system. After saving the first policy, all rollout data from the pre-start stage are discarded, and the RL training proceeds with asynchronous online updates, saving a new policy every $N_2 (N_2 < N_1)$ training steps. Each saved policy will be better. When saving the weights of the $V$-th policy, the trainer worker utilizes the rollout data generated by the $(V-1)$-th policy. 

\noindent\textbf{Policy Update Frequency Trade-Off.} Given $T$ trainer GPUs (batch size $B$ for each GPU) and a replay buffer of size $S_2$, the range of $\{N_1, N_2\}$ is controlled by
\begin{equation}
\begin{cases}
\sigma_{min} \leq \frac{N_2*T*B}{S_2} = \frac{N_1*T*B}{S_1} \leq \sigma_{max}, \\
\quad S_1 > S_2 \geq \sigma.
\end{cases}
\end{equation}
We recommend $\sigma_{min}$ is 8, and $\sigma_{max}$ is 64. Due to the limitations of the 3D mesh generation pre-trained model's capability to learn from rollout data and the complexity of the reward distributions, $\sigma$ should not be set too small, so that the model can adequately learn the reward distributions in the current rollout replay buffer.

\noindent\textbf{Truncated Training.} Based on the truncated training mode in Hourglass Transformer architecture, asynchronous RL generates each complete mesh and then randomly truncates it $\mathcal{T}$ times. Each truncated mesh $M_t$, input point cloud $P$, and corresponding reward $R_t$ form a rollout data ($M_t, P, R_t$). Trainer workers collect rollout data in batches and perform policy updates using RL loss. 

% Inspired by Meshtron \cite{hao2024meshtron}, the face count is predicted by feeding the Vecset \cite{zhang20233dshape2vecset, zhao2023michelangelo} features into a multilayer perceptron (MLP).

%%%%%%%%%%%%%%%%%%%%%%%%%%%%%%%%%%%%%%%%%%%%%%%%%%%%%%%%%%%%%%%%%%%%%%%%%%%%
\subsection{ARPO}
\label{sec:ARPO}

\noindent\textbf{Slow Convergence on Explicit Advantage Modeling.}
% Methods such as GRPO, explicitly model the advantage function distribution. Given a regularization term $R(\pi_\theta, \pi_{\text{ref}})$ (e.g., KL divergence) and advantages, the optimization target of on-policy GRPO series is
% \begin{equation}
% \mathcal{T}_{\text{GRPO}}(\theta) = \hat{\mathbb{E}}_{k}[\hat{A}^{(k)}] = \hat{\mathbb{E}}_{k} [\frac{\pi_\theta}{\text{detach}(\pi_\theta)} \hat{A}^{(k)} - R(\pi_\theta, \pi_{\text{ref}})]
% \end{equation}
% where $\text{detach}(\pi_\theta)$ means that $\pi_\theta$ is separated from the computation graph and does not participate in gradient calculation. 
Methods such as GRPO directly maximize the expectation of the advantage function as the optimization objective, demonstrating stronger generalization capability than implicit reward modeling methods (e.g., DPO). Nevertheless, constrained by the limited capability of pretrained 3D generative models and the inherent complexity of reward distributions, such methods often suffer from low exploration–exploitation efficiency and slow convergence, limiting mesh generation quality (see \cref{sec:Effectiveness_of_ARPO}).

\noindent\textbf{ARPO: Better Trade-off on Training Efficiency and Generalization.}
ARPO samples a group of rollout data and, based on a ranking-based optimization strategy, explicitly leverages advantage values as a weighting mechanism to guide the model in learning the underlying reward distribution. Inspired by \cite{DPO}, starting from a KL-regularized preference optimization objective, the implicit reward is
\begin{equation}
r(x,y)=\beta\log\frac{\pi_\theta(y\mid x)}{\pi_{\mathrm{ref}}(y\mid x)}+\beta\log Z(x), 
\end{equation}
where $\beta$ is a parameter controlling the deviation from the base reference policy $\pi_{\mathrm{ref}}$, $Z(x)$ is a normalization term. For a candidate set $\{y_i\}_{i=1}^K$, the ranking probability follows the Plackett--Luce model \cite{PL_model}
\begin{equation}
p(\omega \mid \{y_i\}_{i=1}^K) = \prod_{k=1}^K \frac{\exp\bigl(\beta\log\frac{\pi_\theta(y_{\omega(k)}\mid x)}{\pi_{\mathrm{ref}}(y_{\omega(k)}\mid x)}\bigr)}{\sum_{j=k}^K \exp\bigl(\beta\log\frac{\pi_\theta(y_{\omega(j)}\mid x)}{\pi_{\mathrm{ref}}(y_{\omega(j)}\mid x)}\bigr)}.
\label{eq:plackett}
\end{equation}
Given rewards $\{R^{(i)}\}_{i=1}^K$ (smaller subscript means stronger preference), the corresponding advantages are calculated as
\begin{equation}
A^{(k)} = \frac{R^{(k)} - \min\{\{R^{(i)}\}_{i=1}^K\} }{\sum_{k=1}^K (R^{(k)} - \min\{\{R^{(i)}\}_{i=1}^K\}) + \epsilon},
\label{eq:advantage}
\end{equation}
where $\epsilon$ is a small constant to avoid division by zero. ARPO minimizes the advantage-weighted negative preference.
\begin{equation}
\mathcal{L}_{\mathrm{ARPO}}
= -\,\mathbb{E}\Bigg[\sum_{i=1}^K A^{(i)}
\log \underbrace{\frac{\exp\!\Big(\beta\log \frac{\pi_\theta(y_i\mid x)}{\pi_{\mathrm{ref}}(y_i\mid x)}\Big)}
{\sum_{j=i}^K \exp\!\Big(\beta\log\frac{\pi_\theta(y_j\mid x)}{\pi_{\mathrm{ref}}(y_j\mid x)}\Big)}}_{\text{predicted probability}}\Bigg],
\end{equation}
Furthermore, the gradient of ARPO is derived as 
\begin{equation}
\begin{aligned}
\nabla_\theta \mathcal{L}_{\text{ARPO}} &= -\beta \cdot \mathbb{E} \left[ \sum_{i=1}^{K-1} \sum_{j=i+1}^K A^{(i)} \cdot \sigma\left( -\beta \cdot \Delta_{i,j} \right) \cdot \right. \\
& \quad \left. \left( \frac{\nabla_\theta \pi_\theta(y_i|x)}{\pi_\theta(y_i|x)} - \frac{\nabla_\theta \pi_\theta(y_j|x)}{\pi_\theta(y_j|x)} \right) \right],
\end{aligned}
\end{equation}
where preference difference is 
\begin{equation}
\Delta_{i,j} = \log \frac{\pi_\theta(y_i|x)}{\pi_{\text{ref}}(y_i|x)} - \log \frac{\pi_\theta(y_j|x)}{\pi_{\text{ref}}(y_j|x)}.
\end{equation}

High-dominance samples are given greater weight, and the model focuses more on ``high-value'' preference pairs. Low-dominance samples have reduced weight to reduce the impact of noise. ARPO can be seen as the maximization of the advantage function under the ``predicted probability'' \cite{NCA} of an implicit reward representation. ARPO is essentially a ranking strategy model with fast and stable convergence. Also, the explicit advantage function guidance enables a deeper understanding and modeling of reward distributions, demonstrating stronger generalization capabilities. ARPO achieves a better trade-off between training efficiency and generalization than existing RL algorithms designed for 3D mesh generation, such as commonly used DPO and GRPO. 

\noindent\textbf{ARPO in Asynchronous RL Framework.}
ARPO is seamlessly integrated into our asynchronous online RL framework. Each rollout worker samples a group of samples and stores it in the rollout replay buffer, and then the trainer worker uses ARPO loss to learn a better policy. This asynchronous design ensures that the ARPO objective is optimized on a relatively timely dataset of mesh generations, leading to more efficient training compared to synchronous methods and offline methods. 

\noindent\textbf{Truncated Asynchronous ARPO Training.}
To handle the varying lengths of 3D meshes and improve training efficiency, we employ truncated training within the asynchronous ARPO framework. Trainer workers process collected rollout data $(M_t, P, R_t)$, calculate $A_t$ for each $M_t$, and then perform policy learning with the ARPO objective for truncated meshes. This approach ensures that the model learns from a wider range of mesh structures. Truncated ARPO loss is represented as 
\begin{equation}
\mathcal{L}_{\mathrm{t-ARPO}}
= -\,\mathbb{E}_{M_t}\Bigg[\sum_{i=1}^K A_t^{(i)}
\log\frac{\exp\!\Big(\mathcal{R}_{i|m:m+w}\Big)}
{\sum_{j=i}^K \exp\!\Big(\mathcal{R}_{j|m:m+w}\Big)}\Bigg], 
\end{equation} 
\begin{equation}
\mathcal{R}_{i|m:m+w} = \beta \log\frac{\pi_\theta(y_{i|m:m+w}\mid P)}{\pi_{\mathrm{ref}}(y_{i|m:m+w}\mid P)},
\end{equation} 
where $m$ and $w$ are the initial position and truncated window length (36,864 tokens), respectively. Truncated asynchronous ARPO training teaches RL policy to make locally optimal decisions that lead to globally superior topology.

\begin{figure*}[t]
    \centering
    \includegraphics[width=1.0\textwidth]{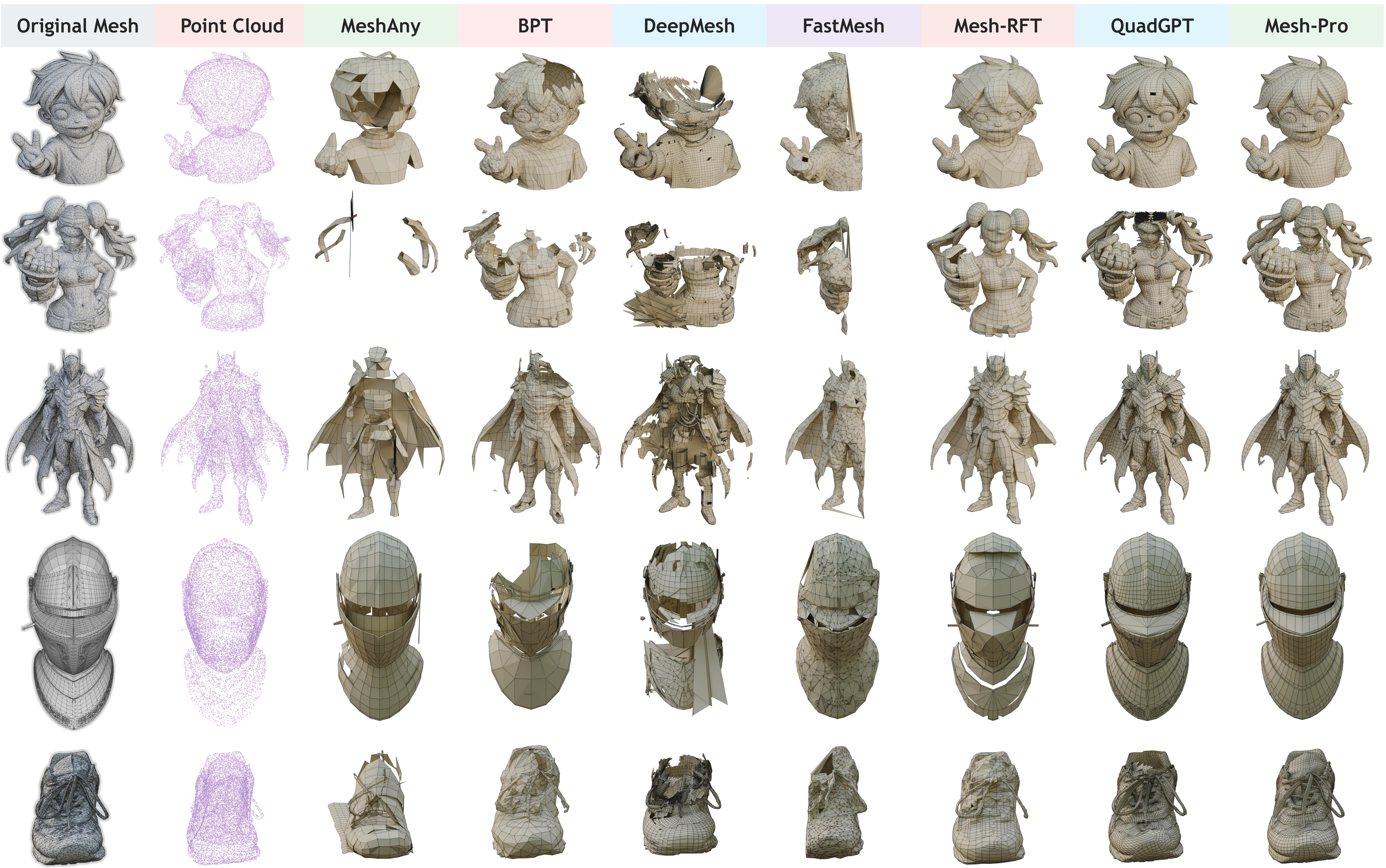}
    \caption{ 
    Qualitative comparison of Mesh-Pro with other methods.
    }
    \label{fig:qualitative_comparison}
\end{figure*}

\subsection{Reward Design}
\label{sec:reward}
A robust reward function is essential for guiding the mesh generation process toward high-quality outputs that align with artistic styles. Our reward model captures multiple facets of mesh quality, including topological regularity, structural integrity, and aesthetic appeal. The reward function $R(M_t)$ for a generated mesh $M_t$ given a prompt $P$ is 
\begin{align}
\scalebox{0.93}{$
R(M_t) = \begin{cases} w_{\text{qr}} \cdot N_{\text{qr}} + N^2_{\text{ql}} & \text{if } N_\text{bf} < \theta_{ray} \text{ and } D_\text{hd} < \theta_{hd}, \\ 0 & \text{otherwise}. \end{cases}
$}
\end{align}
\noindent\textbf{Ray Casting Integrity Reward (\boldmath $R_{\text{ray}}(M_t)$).}
Rays are cast toward the mesh from multiple directions. Mesh quality is assessed by comparing the normal direction of the surface hit by each ray with the ray's direction. Incomplete or inconsistent regions, such as broken surfaces, result in the identification of bad faces. If the number of bad faces $N_\text{bf}$ exceeds a predefined threshold $\theta_{ray}$, then $R(M_t)$ is set to 0. Further details are provided in the supplementary materials.

\noindent\textbf{Topological Reward (\boldmath $R_{\text{topo}}(M_t)$).}
Inspired by \cite{liu2025quadgpt}, we emphasize structured edge flow as a hallmark of high-quality quad topology. This is quantified by identifying two key structures: quad rings (closed loops of faces) and quad lines (open strips of faces). We employ an edge-based traversal that moves across adjacent quadrilateral faces. Paths that terminate are classified as quad lines, while those that return to the starting edge form quad rings. The topological reward $R_{\text{topo}}(M_t)$ is computed as the weighted sum of the number of quad rings $N_{\text{qr}}$ and the square of the number of quad lines $N^2_{\text{ql}}$ within the analyzed region.

\noindent\textbf{Hausdorff Distance Consistency Reward (\boldmath $R_{\text{hd}}(M_t)$).}
To ensure geometric fidelity between the input point cloud and the generated mesh, we employ the Hausdorff distance $D_\text{hd}$. If $D_\text{hd} \geq \theta_{hd}$, then $R(M_t)$ is set to 0.

The hyperparameter $w_{\text{qr}}$ balances the contributions of different topological aspects. This multifaceted reward function enables a comprehensive evaluation of generated meshes, steering the model toward outputs that are not only geometrically complete and consistent but also topologically regular and aesthetically pleasing.

\section{Experiments}

\subsection{Dataset}
The pre-training of Mesh-Pro utilizes a dataset of 1.3 million quad-dominant meshes, which we assemble and curate through a multi-stage process. This process involves collecting, converting, and filtering data from various sources, including Objaverse~\citep{deitke2023objaverse}, Objaverse-XL~\citep{deitke2023objaversexl}, ShapeNetV2~\citep{chang2015shapenet}, and 3D-FUTURE~\citep{fu20213d}, alongside privately licensed assets. For asynchronous ARPO post training, we use 500 diverse high-fidelity meshes generated from Hunyuan3D 2.5 \citep{lai2025hunyuan3d} and 200 artist meshes. See supplementary materials for more details.

\subsection{Implementation Details}
For pre-training, Mesh-Pro is trained on a cluster of 64 NVIDIA H20 GPUs for 7 days, with a learning rate of 1e-4. The truncated mesh decoder has 1.1B parameters. Point clouds are densely sampled at 40,960 points, and the point cloud encoder is built upon Michelangelo \cite{zhao2023michelangelo}. Asynchronous ARPO is trained for 1 days using 64 GPUs for rollout and 32 GPUs for policy updates, with a learning rate of 1e-6. The default group is 4, and $\beta$ is 1.0. $N_1$ and $\sigma$ are set to 2000 and 1000, respectively. Truncated training randomly performs truncation $\mathcal{T}=4$ times. The reward weight $w_{\text{qr}}$ is 0.1. $\theta_{ray}$ and $\theta_{hd}$ is 1 and 0.1, respectively. On the NVIDIA H20 GPU, with KV caching and CUDA graph acceleration, the inference speed of Mesh-Pro is approximately \textbf{310} token/s. 

\subsection{Baselines and Evaluation Metrics}
Since current methods are mainly designed to produce triangular outputs, we apply a robust triangle-to-quadrilateral conversion algorithm \cite{remacle2012blossom, liu2025quadgpt} as a post-processing step to facilitate a fair comparison. For methods that have not been open-sourced, we reproduce their implementations. Due to significant GPU idle time and training interruptions encountered by synchronous RL when training meshes of varying lengths, it is excluded from further experimental comparisons. Our primary comparison is therefore with existing RL methods utilized in 3D mesh generation.

We report metrics for geometric fidelity (CD, HD), geometric integrity (Broken Ratio, BR), and topological quality (Quad Ratio, QR), alongside a comprehensive User Study (US) to assess perceptual quality. Human experts rank the generated meshes of $U$ methods from best ($U-1$) to worst ($0$), revealing relative preference relationships. We use US as the primary subjective metric for mesh generation.

\begin{figure}[t]
    \centering
    \includegraphics[width=0.4\textwidth]{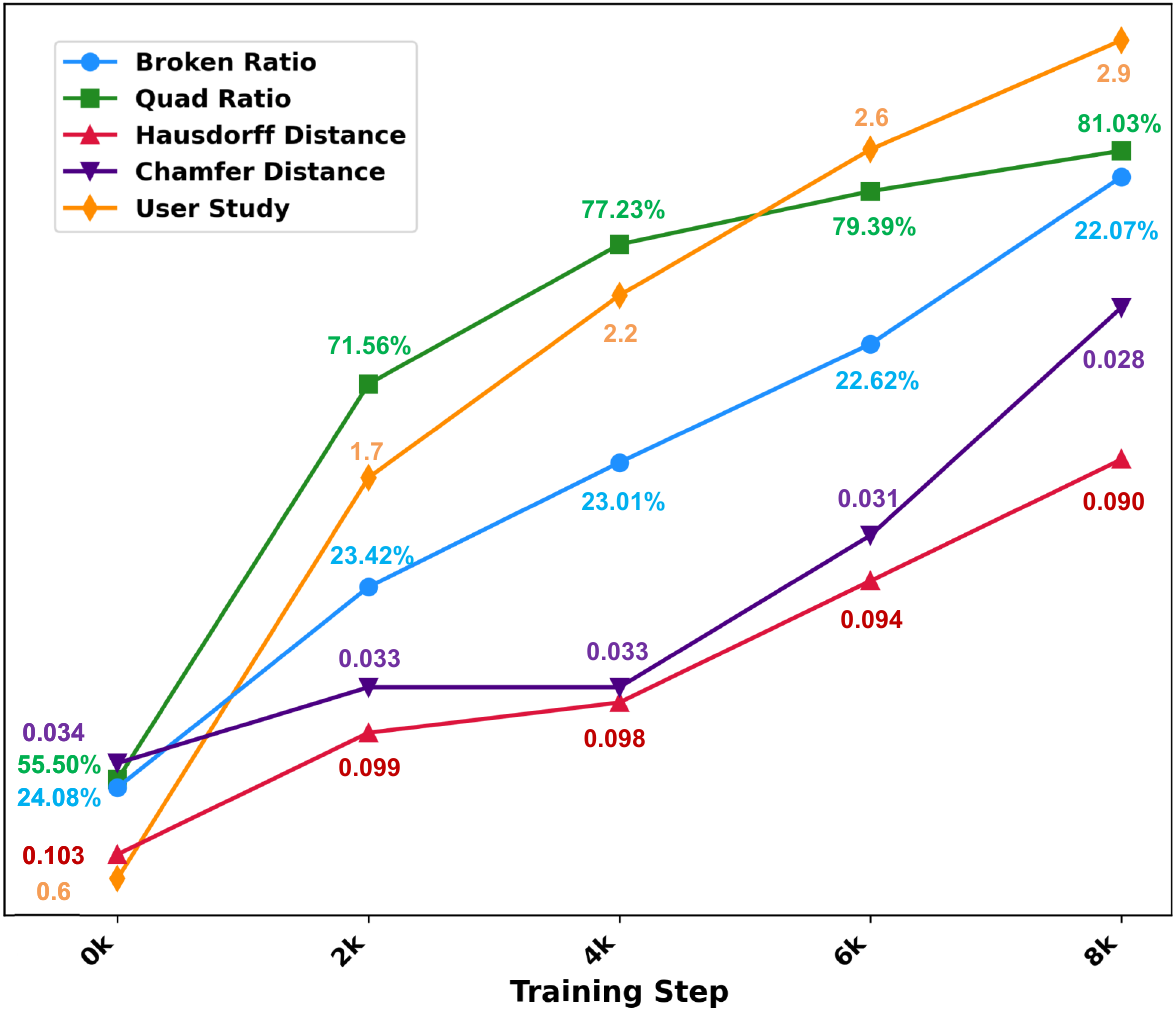}
    \caption{ 
    Performance of asynchronous ARPO over training steps. Y-axes for ``$\downarrow$'' metrics (BR, CD, HD) are intentionally inverted to align upward trends with performance gains (similarly below).
    }
    \label{fig:Async_ARPO_curve}
\vspace{-0.2cm}
\end{figure}

\begin{figure}[t]
    \centering
    \includegraphics[width=0.48\textwidth]{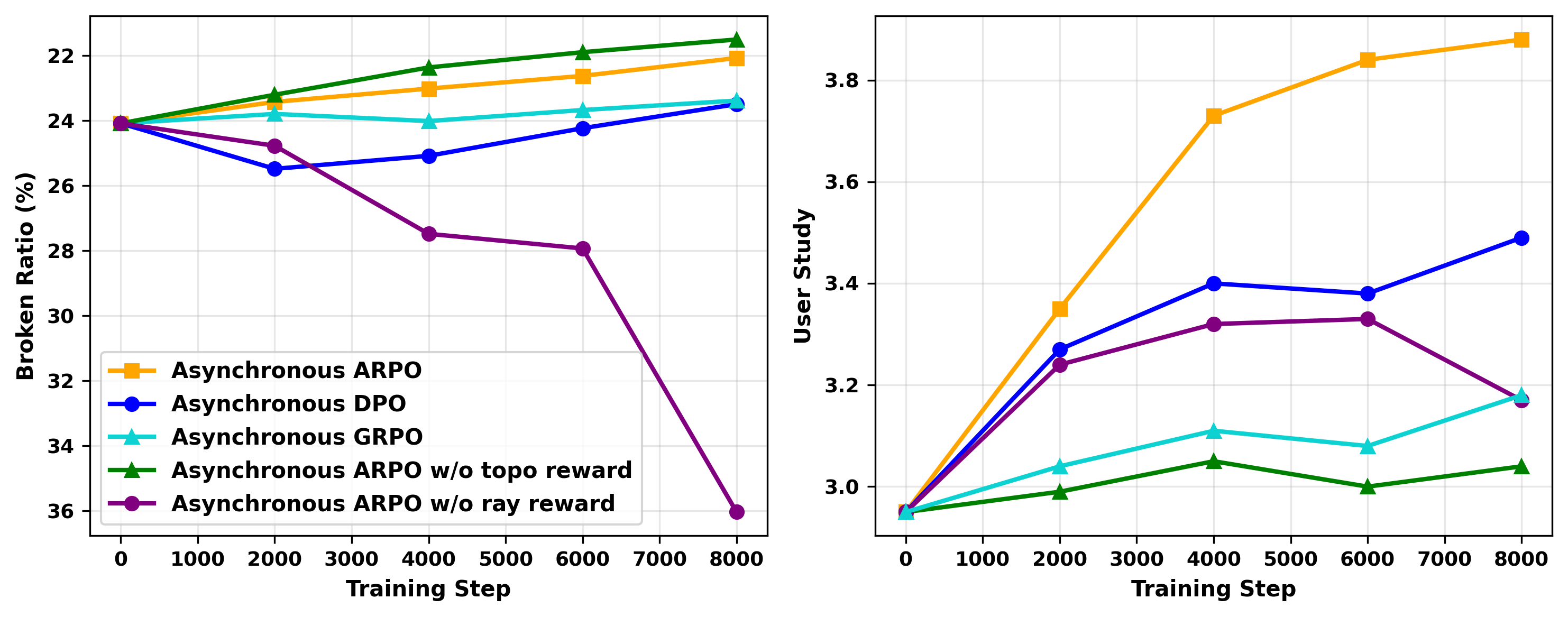}
    \caption{ 
    Performance curves of asynchronous RL methods and reward function over training steps.
    }
    \label{fig:Async_RL_curve}
\vspace{-0.2cm}
\end{figure}

\subsection{Qualitative Results}
We present a qualitative comparison against existing mesh generation methods on two challenging domains: out-of-distribution meshes from other generative models and in-distribution, high-quality artist-created assets. As illustrated in \cref{fig:qualitative_comparison}, current methods often fail to produce viable geometry. They frequently yield meshes with severe topological artifacts, including broken geometry, irregular topology, holes, and fragmented surfaces. In stark contrast, Mesh-Pro consistently generates topologically sound and geometrically faithful reconstructions across all examples.

\subsection{Quantitative Results}
In quantitative comparison in \cref{tab:quantitative_comparison}, dense meshes are selected from Hunyuan3D 2.5, while artist meshes are taken from the public Toys4k \cite{Toys4k} dataset. Mesh-Pro sets a new performance benchmark on both dense and artist-designed meshes, outperforming prior works by a significant margin. This advantage is driven by our asynchronous ARPO, quadrilateral tokenization, and reward function, which synergistically optimizes for both geometric and topological quality. The ray-based reward $R_{\text{ray}}$ ensures structural integrity, leading to the lowest broken ratios, while the topological reward $R_{\text{topo}}$ fosters clean, structured quad layouts. Mesh-Pro achieves the highest US scores, confirming that Mesh-Pro more closely approaches artist-level standards.

\begin{table}
\centering
\setlength{\tabcolsep}{3pt}
\caption{Quantitative comparison on Dense and Artist Meshes.}
\resizebox{1.05\columnwidth}{!}{
    \begin{tabular}{@{}l|ccccc|ccccc@{}}
    \toprule  
    Data Type &\multicolumn{5}{c|}{Dense Meshes} & \multicolumn{5}{c@{}}{Artist Meshes} \\
    \midrule  
    Metrics & CD $\downarrow$ & HD $\downarrow$ & BR $\downarrow$ & QR $\uparrow$ & US $\uparrow$ & CD $\downarrow$ & HD $\downarrow$ & BR $\downarrow$ & QR $\uparrow$  &US $\uparrow$ \\
    \midrule  
    MeshAnyv2~\citep{chen2024meshanythingv2} & 0.148 & 0.395 & 94\% & 54\% & 1.5 & 0.084 & 0.239 & 95\% & 57\% & 1.8 \\
    BPT~\citep{weng2025scaling}                 & 0.109 & 0.284 & 72\% & 43\% & 2.5 & 0.046 & 0.131 & 50\% & 47\% & 2.5\\
    DeepMesh~\citep{zhao2025deepmesh}           & 0.351 & 0.635  & 91\% & 64\% & 2.8 & 0.336 & 0.584 & 64\% & 66\% & 2.7\\
    FastMesh~\citep{kim2025fastmesh}            & 0.099 & 0.256 & 99\% & 4\% & 1.1 & 0.044 & 0.133 & 98\% & 18\% & 1.2 \\
    Mesh-RFT~\citep{liu2025meshrft}            & 0.051 & 0.161 & 40\% & 65\% & 3.4 & 0.041 & 0.134 & 38\% & 69\% & 3.5 \\
    QuadGPT~\citep{liu2025quadgpt}      & 0.059 & 0.163 & 50\% & 78\% & 4.5 & 0.042 & 0.101 & 39\% & 76\% & 4.4 \\
    Mesh-Pro         & \textbf{0.028} & \textbf{0.090} & \textbf{22\%} & \textbf{81\%} & \textbf{5.2} & \textbf{0.038} & \textbf{0.094} & \textbf{32\%} & \textbf{78\%} & \textbf{4.9} \\
    \bottomrule  
    \end{tabular}
}
\label{tab:quantitative_comparison}
\end{table}

% \subsection{Runtime Analysis}

% \begin{table}[t]
% \centering
% \setlength{\tabcolsep}{3pt}
% \caption{Runtime Analysis of Mesh-Pro.}
% \resizebox{0.8\columnwidth}{!}{
%     \begin{tabular}{@{}lccc@{}}
%     \toprule  
%     Configuration & Baseline & +KV caching & + CUDA graph \\ 
%     \midrule  
%     Token/s $\uparrow$ & 40 & 110 & 310 \\
%     \bottomrule  
%     \end{tabular}
% }
% \label{tab:Runtime}
% \end{table}

% \cref{{tab:Runtime}} presents the inference time analysis of Mesh-Pro. On the NVIDIA H20 GPU, with KV caching and CUDA graph acceleration, the inference speed of Mesh-Pro is approximately 310 token/s. 

\begin{table}[t]
\centering
\setlength{\tabcolsep}{3pt}
\caption{Effectiveness analysis of each component of asynchronous ARPO. ``*'' denotes the pretrained model.}
\resizebox{1.0\columnwidth}{!}{
    \begin{tabular}{@{}cccc|ccccc@{}}
    \toprule  
    \makecell{Group} & \makecell{Advantage} & \makecell{Online} & \makecell{Pre-Start} & CD $\downarrow$ & HD $\downarrow$ & BR $\downarrow$ & QR $\uparrow$ & US $\uparrow$ \\ 
    \midrule  
    * & * & * & *        & 0.034 & 0.103 & 24\% & 55\% & 1.4 \\
    2 & & &               & 0.032 & 0.102 & 23\% & 70\% & 2.1 \\
    4 & & &              & 0.030 & 0.097 & 24\% & 73\% & 2.4 \\
    4 & \ding{51} & &     & 0.029 & 0.093 & 23\% & 76\% & 2.9 \\
    4 & \ding{51} & \ding{51} & & 0.033 & 0.103 & 23\% & 74\% & 2.7 \\
    4 & \ding{51} &  \ding{51}  & \ding{51}  & \textbf{0.028} & \textbf{0.090} & \textbf{22\%} & \textbf{81\%} & \textbf{3.5} \\
    \bottomrule  
    \end{tabular}
}
\label{tab:analysis_Async_RL}
\end{table}

\subsection{Ablation Study}
\subsubsection{Asynchronous Online RL}
We conduct an analysis of synchronous and asynchronous RL training under the same computational hardware conditions. A comparison of the training time for the same number of steps on an equivalent number of meshes reveals that our proposed asynchronous online RL is approximately \textbf{3.75$\times$} faster than synchronous RL. 

\cref{fig:Async_ARPO_curve} illustrates that the performance metrics of the mesh generated by Mesh-Pro consistently and steadily improve as the asynchronous ARPO training progresses. In \cref{tab:analysis_Async_RL}, asynchronous online RL achieves better performance than offline RL on all metrics. By introducing the Pre-Start stage, asynchronous online ARPO training becomes more stable, leading to higher-quality meshes.

\begin{figure*}[t]
    \centering
    \includegraphics[width=0.95\textwidth]{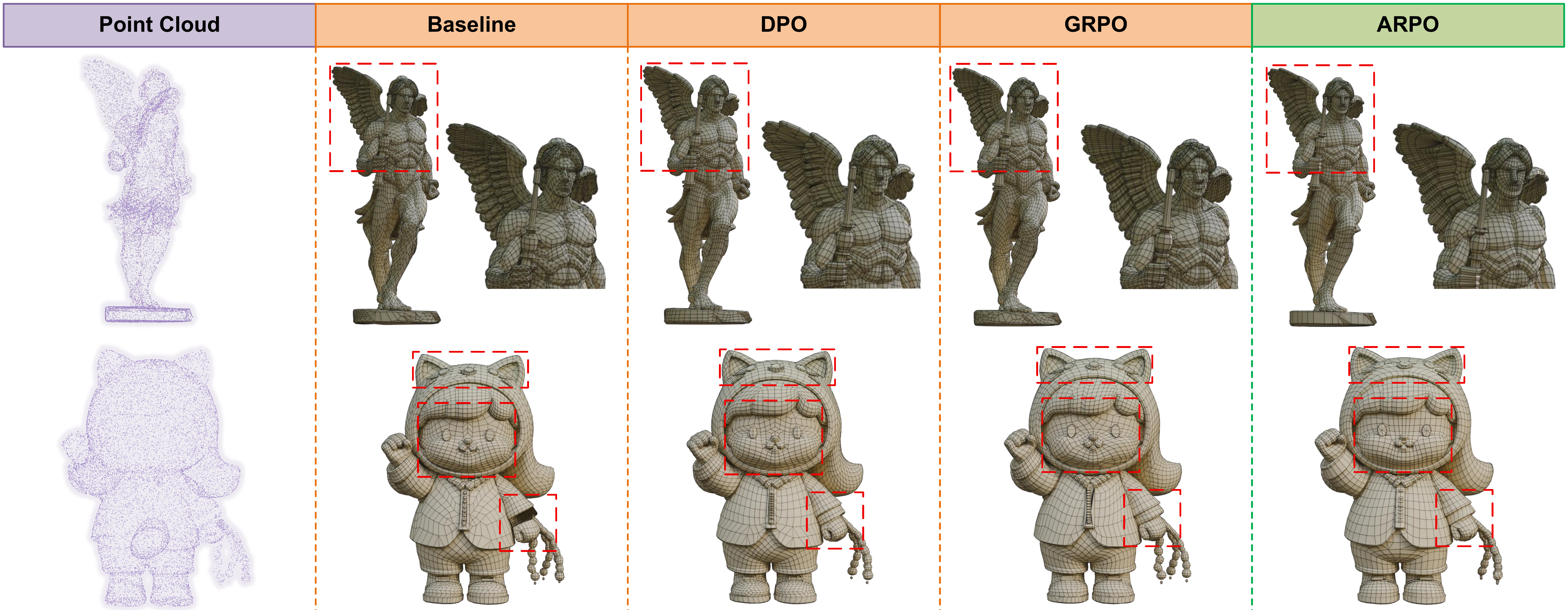}
    \caption{
    Visual comparison of different asynchronous online RL algorithms.
    }
    \label{fig:RL_comparison}
\vspace{-0.2cm}
\end{figure*}

\subsubsection{Effectiveness of ARPO}
\label{sec:Effectiveness_of_ARPO}
We first analyze the components of ARPO. In \cref{tab:analysis_Async_RL} and \cref{fig:Async_RL_curve}, the explicit advantage guidance improves the model's generalization ability and the quality of the generated meshes, while maintaining stable and fast convergence. Compared to the offline ARPO, the asynchronous online ARPO continuously enhances the geometric consistency, integrity, and topological quality.

Furthermore, under identical settings, we evaluate DPO used in prior RL-based mesh generation, and explicit reward modeling RL methods such as GRPO, as presented in \cref{tab:ARPO_DPO_GRPO}. \cref{fig:Async_RL_curve} gives the performance curves of asynchronous RL over training steps. To facilitate multi-dimensional comparisons, US in \cref{fig:Async_RL_curve} is based on absolute quality scores 1-5. In comparison, ARPO achieves a better trade-off between training efficiency and generalization performance. ARPO exhibits superior generalization performance over DPO, while also achieving faster and more stable convergence compared to GRPO. The quality of the meshes generated by ARPO is significantly superior to that of DPO and GRPO. GRPO suffers from inefficient exploration-exploitation, low training efficiency, and slow performance improvement. \cref{fig:RL_comparison} provides a visual comparison of the mesh quality before and after RL using ARPO, alongside the results from DPO and GRPO.

\begin{table}[t]
\centering
\setlength{\tabcolsep}{3pt}
\caption{Quantitative comparison of ARPO with DPO and GRPO.}
\resizebox{0.8\columnwidth}{!}{
    \begin{tabular}{@{}l|ccccc@{}}
    \toprule  
    Metrics & CD $\downarrow$ & HD $\downarrow$ & BR $\downarrow$ & QR $\uparrow$ & US $\uparrow$ \\
    \midrule  
    Pretrained Model              & 0.034 & 0.103 & 24\% & 55\% & 0.4 \\
    \midrule
    with Async DPO   & \textbf{0.028} & 0.093 & 23\% & 75\% & 1.9 \\
    with Async GRPO       & 0.033 & 0.096 & 23\% & 68\% & 1.3 \\
    with Async ARPO       & \textbf{0.028} & \textbf{0.090} & \textbf{22\%} & \textbf{81\%} & \textbf{2.4} \\
    \bottomrule  
    \end{tabular}
}
\label{tab:ARPO_DPO_GRPO}
\vspace{-0.2cm}
\end{table}

\begin{table}[t]
\centering
\setlength{\tabcolsep}{3pt}
\caption{Performance with scaled data in asynchronous ARPO.}
\resizebox{0.42\columnwidth}{!}{
    \begin{tabular}{@{}cccc@{}}
    \toprule  
    Data Scale & BR $\downarrow$  & US $\uparrow$\\ 
    \midrule  
    400  & 22.35\% & 0.86 \\
    700  & 22.07\% & 1.01 \\
    1000 & \textbf{21.83\%} & \textbf{1.13} \\
    \bottomrule  
    \end{tabular}
}
\label{tab:Data_Scaling}
\vspace{-0.2cm}
\end{table}

\begin{table}[t]
\centering
\setlength{\tabcolsep}{3pt}
\caption{Comparison of different mesh tokenizations.}
\resizebox{0.87\columnwidth}{!}{
    \begin{tabular}{@{}c|c|ccccc@{}}
    \toprule  
    Tokenizer & Async ARPO & CD $\downarrow$ & HD $\downarrow$ & BR $\downarrow$ & QR $\uparrow$ & US $\uparrow$ \\
    \midrule  
    \multirow{2}{*}{Previous} & w/o & 0.056 & 0.122 & 38\% & 72\% & 1.2 \\
                              & with & 0.071 & 0.160 & 46\% & 80\% & 1.8 \\
    \midrule
    \multirow{2}{*}{Ours} & w/o & 0.034 & 0.103 & 24\% & 55\% & 0.9 \\
                          & with & \textbf{0.028} & \textbf{0.090} & \textbf{22\%} & \textbf{81\%} & \textbf{2.1} \\
    \bottomrule  
    \end{tabular}
}
\label{tab:mesh_tokenization}
\vspace{-0.4cm}
\end{table}

\subsubsection{Impact of Data Scale in Asynchronous ARPO}
The impact of training data scale on asynchronous ARPO performance is evaluated in \cref{tab:Data_Scaling}. Larger data scales improve performance, revealing its data scaling potential.

\subsubsection{Mesh tokenization}
\cref{tab:mesh_tokenization} presents a comparison of different tokenization methods. Our tokenization generates more stable meshes with a significantly lower structural broken ratio than prior works. Furthermore, after applying asynchronous ARPO, our tokenization demonstrates superior geometric consistency, structural integrity, and topological soundness.

\subsubsection{Reward Design}
\cref{fig:Async_RL_curve} confirms the effectiveness of our reward components. The absence of the ray reward $R_{\text{ray}}$ causes a substantial rise in the mesh broken ratio. Removing the topology reward $R_{\text{topo}}$ diminishes the output quality, moving it further from an artist-level benchmark. Since ensuring mesh integrity takes precedence over topology elegance, our ray reward is crucial for generating high-quality quad-meshes.

% \begin{table}
% \centering
% \setlength{\tabcolsep}{3pt}
% \caption{Ablation Study 1.}
% \resizebox{1.05\columnwidth}{!}{
%     \begin{tabular}{@{}cccc|ccccc@{}}
%     \toprule  
%     & \makecell{Group\\Number} & \makecell{Advantage\\Guidance} & \makecell{Asynchronous\\Online} & CD $\downarrow$ & HD $\downarrow$ & BR $\downarrow$ & QR $\uparrow$ & US $\uparrow$ \\ 
%     \midrule  
%     Baseline & -- & -- & --         & 0.034 & 0.103 & 24\% & 55\% &  \\
%     Offline DPO(BT) & 2 & &                &  &  &  &  &  \\
%     Offline DPO(PL) & 4 & &                &  &  &  &  &  \\
%     Offline ARPO & 4 & \ding{51} &      & 0.032 & 0.101 & 23\% & 73\% &  \\
%     Async Online ARPO & 4 & \ding{51} & \ding{51}  & \textbf{0.028} & \textbf{0.090} & \textbf{22\%} & \textbf{81\%} & \\
%     \bottomrule  
%     \end{tabular}
% }
% \label{tab:Ablation_Study_1}
% \end{table}

\section{Conclusion}

This paper proposes an asynchronous online ARPO that demonstrates significant advantages in RL post-training efficiency and generalization performance over offline DPO, which has been predominantly relied upon in existing 3D mesh generation works. Built upon asynchronous ARPO, we propose Mesh-Pro, which integrates a novel diagonal-aware mixed triangular-quadrilateral tokenization and a ray-based reward for geometric integrity, achieving SOTA performance in artist-style quadrilateral mesh generation. We hope to inspire RL-driven 3D generation toward transformative advances akin to RL-driven text generation.

% \section*{Acknowledgments} 
% % This research was supported by fundings from the Hong Kong RGC General Research Fund (152228/23E, 162161/24E, 162116/25E, 162180/25E), National Natural Science Foundation of China (NSFC) Key Program (No.62532005), Collaborative Research Fund (No. C1042-23GF, No. 5097-25G), NSFC/RGC Collaborative Research Scheme (Grant No. 62461160332 \& CRS\_HKUST602/24), Research Impact Fund (No. R5011-23F), Areas of Excellence Scheme (AoE/E-601/22-R), and the InnoHK (HKGAI).

% This work was supported in part by the National Natural Science Foundation of China (NSFC) under Grants 62573413 and 62503475, and the NSFC Key Program under Grant 62532005; the National Key Research and Development Program of China under Grant 2023YFB4706800; the Key Research and Development Program of Sichuan Province under Grant 2024YFCY0029-BC-05; the Hong Kong RGC General Research Fund (152228/23E, 162161/24E, 162116/25E, 162180/25E); the Collaborative Research Fund (No.~C1042-23GF, No.~5097-25G); the NSFC/RGC Collaborative Research Scheme (Grant No.~62461160332 \& CRS\_HKUST602/24); the Research Impact Fund (No.~R5011-23F); the Areas of Excellence Scheme (AoE/E-601/22-R); and the InnoHK (HKGAI).
{
    \small
    \bibliographystyle{ieeenat_fullname}
    \bibliography{main}
}

\clearpage
\setcounter{page}{1}
\maketitlesupplementary

\appendix

% 恢复目录深度，开始记录补充材料的章节
\addtocontents{toc}{\protect\setcounter{tocdepth}{3}}

\renewcommand{\contentsname}{Table of Contents} 
\tableofcontents

\clearpage

\section{Mesh-Pro}
\subsection{Diagonal-Aware Mesh Tokenization}
\label{sec:sup_tokenizer}

The procedure of diagonal-aware mesh tokenization is outlined in \cref{alg:diag-tokenizer}. The proposed tokenization enhances prior approaches \cite{liu2025quadgpt} by incorporating a canonical global order and a strategy that prioritizes the generation of triangular faces before selectively forming quadrilateral faces. 

\RestyleAlgo{ruled}
\begin{algorithm}
\caption{Diagonal-Aware Tokenization}\label{alg:diag-tokenizer}
\SetStartEndCondition{ }{}{}%
\SetKwProg{Fn}{def}{\string:}{}
\SetKw{KwTo}{in}\SetKwFor{For}{for}{\string:}{}%
\SetKwIF{If}{ElseIf}{Else}{if}{:}{elif}{else:}{}%
\SetKwComment{Comment}{/* }{ */}
\AlgoDontDisplayBlockMarkers\SetAlgoNoEnd\SetAlgoNoLine%

\KwData{Discretized Vertices $\mathbf V = \{\mathbf{v}_i\}_N$, Mixed Faces $\mathbf F = \{\mathbf f_i\}_M$ (triangles and quads, with both vertices and faces already ordered by previous serialization~\cite{liu2025quadgpt}).}
\KwResult{Token sequence $\mathbf O$, Special token $S = 2^{n_{bits}}$.}
\BlankLine

\SetKwFunction{SerializeFace}{SerializeFace}%
\SetKwFunction{WriteVertex}{WriteVertex}%

\Fn(){\SerializeFace{Face $\mathbf{f}$}}{
    \uIf{$|\mathbf{f}| = 3$}{
        \tcc{Triangle: pad with $3S$}
        \For{$\mathbf{v}$ \KwTo $\mathbf{f}$}{
            \WriteVertex($\mathbf{v}$)\;
        }
        $\mathbf{O}$.append($\mathbf{v}_1.x,\mathbf{v}_1.y,\mathbf{v}_1.z,\mathbf{v}_2.x,\mathbf{v}_2.y,\mathbf{v}_2.z$,
        $\mathbf{v}_3.x,\mathbf{v}_3.y,\mathbf{v}_3.z$, $3S, 3S, 3S$)\;
    }
    \ElseIf{$|\mathbf{f}| = 4$}{
        \tcc{Quad ABCD (diagonal AC, A$<$C): min vertex first, smaller triangle (ABC or ACD) first}
        $i_{\min} \gets \arg\min_i \mathbf{f}[i]$\;
        \uIf{$i_{\min} = 0$ \KwSty{and} $\mathbf{f}[1] < \mathbf{f}[3]$}{
            $\mathbf{f}' \gets [\mathbf{f}[0], \mathbf{f}[1], \mathbf{f}[2], \mathbf{f}[3]]$; $flag \gets 0$\;
        }
        \uElseIf{$i_{\min} = 0$ \KwSty{and} $\mathbf{f}[1] \geq \mathbf{f}[3]$}{
            $\mathbf{f}' \gets [\mathbf{f}[0], \mathbf{f}[2], \mathbf{f}[3], \mathbf{f}[1]]$; $flag \gets 1$\;
        }
        \uElseIf{$i_{\min} = 1$}{
            $\mathbf{f}' \gets [\mathbf{f}[1], \mathbf{f}[2], \mathbf{f}[0], \mathbf{f}[3]]$; $flag \gets 2$\;
        }
        \Else{
            $\mathbf{f}' \gets [\mathbf{f}[3], \mathbf{f}[0], \mathbf{f}[2], \mathbf{f}[1]]$; $flag \gets 2$\;
        }
        
        \For{$j \gets 0$ \KwTo $2$}{
            \WriteVertex($\mathbf{f}'[j]$)\;
        }
        \tcc{Encode diagonal flag in 4th vertex}
        $\mathbf{v}_1 \gets \mathbf{f}'[0]$\;
        $\mathbf{v}_2 \gets \mathbf{f}'[1]$\;
        $\mathbf{v}_3 \gets \mathbf{f}'[2]$\;
        $\mathbf{v}_4 \gets \mathbf{f}'[3]$\;
    $\mathbf{O}$.append($\mathbf{v}_1.x,\mathbf{v}_1.y,\mathbf{v}_1.z,\mathbf{v}_2.x,\mathbf{v}_2.y,\mathbf{v}_2.z$,
    $\mathbf{v}_3.x,\mathbf{v}_3.y,\mathbf{v}_3.z$,
    $\mathbf{v}_4.x + flag \cdot S$, $\mathbf{v}_4.y + flag \cdot S$, $\mathbf{v}_4.z + flag \cdot S$)\;
    }
}

\tcc{Main serialization loop}
\For{Face $\mathbf{f}$ \KwTo $\mathbf{F}$}{
    \SerializeFace($\mathbf{f}$)\;
}

\end{algorithm}

The procedure for reconstructing the mixed triangle-quadrilateral mesh from the generated sequence of tokens is detailed in \cref{alg:diag-detokenizer}. The process yields either triangular or quadrilateral faces, contingent on the presence of padding tokens (``S''). Subsequently, we resolve the diagonal states for quadrilaterals based on the designated flags.

\begin{algorithm}[h]
\caption{Diagonal-Aware Detokenization}\label{alg:diag-detokenizer}
\SetStartEndCondition{ }{}{}%
\SetKwProg{Fn}{def}{\string:}{}
\SetKw{KwTo}{in}\SetKwFor{For}{for}{\string:}{}%
\SetKwIF{If}{ElseIf}{Else}{if}{:}{elif}{else:}{}%
\SetKwComment{Comment}{/* }{ */}
\AlgoDontDisplayBlockMarkers\SetAlgoNoEnd\SetAlgoNoLine%

\KwData{Token sequence $\mathbf O$, Special token $S = 2^{n_{bits}}$.}
\KwResult{Vertices $\mathbf V$, Faces $\mathbf F$.}
\BlankLine

\SetKwFunction{DeserializeFace}{DeserializeFace}%
\SetKwFunction{ReadVertex}{ReadVertex}%
\SetKwFunction{RecoverQuad}{RecoverQuad}%

\Fn(){\DeserializeFace{Tokens $\mathbf{seq}[0:12]$}}{
    \uIf{$\mathbf{seq}[11] = 3S$}{
        \tcc{Triangle case}
        \For{$i \gets 0$ \KwTo $2$}{
            $\mathbf{v}_i \gets$ \ReadVertex($\mathbf{seq}[3i:3i+3]$)\;
        }
        \Return triangle $[\mathbf{v}_0, \mathbf{v}_1, \mathbf{v}_2]$\;
    }
    \Else{
        \tcc{Quad: decode diagonal flag}
        $flag \gets 0$\;
        \For{$i \gets 0$ \KwTo $3$}{
            $\mathbf{xyz} \gets []$\;
            \For{$j \gets 0$ \KwTo $2$}{
                $val \gets \mathbf{seq}[3i + j]$\;
                \uIf{$val \geq 2S$}{
                    $val \gets val - 2S$; $flag \gets 2$\;
                }
                \uElseIf{$val \geq S$}{
                    $val \gets val - S$; $flag \gets 1$\;
                }
                $\mathbf{xyz}$.append($val$)\;
            }
            $\mathbf{v}_i \gets$ \ReadVertex($\mathbf{xyz}$)\;
        }
        
        \tcc{Recover original vertex order}
        \uIf{$flag = 0$}{
            $\mathbf{f} \gets [\mathbf{v}_0, \mathbf{v}_1, \mathbf{v}_2, \mathbf{v}_3]$\;
        }
        \uElseIf{$flag = 1$}{
            $\mathbf{f} \gets [\mathbf{v}_0, \mathbf{v}_3, \mathbf{v}_1, \mathbf{v}_2]$\;
        }
        \Else{
            $\mathbf{f} \gets [\mathbf{v}_2, \mathbf{v}_0, \mathbf{v}_1, \mathbf{v}_3]$\;
        }
        \Return quad $\mathbf{f}$\;
    }
}

\BlankLine
\tcc{Main deserialization loop}
\For{$i \gets 0$ \KwTo $|\mathbf{O}|$ step $12$}{
    $\mathbf{f} \gets$ \DeserializeFace($\mathbf{O}[i:i+12]$)\;
    $\mathbf{F}$.append($\mathbf{f}$)\;
}

\end{algorithm}

The proposed tokenization method leads to more robust mesh generation. Subsequently, trained with Asynchronous ARPO reinforcement learning (RL), Mesh-Pro demonstrates superior performance over prior approaches.

\newpage
\clearpage

\subsection{Python Code of ARPO}
\begin{figure}[t]
    \centering
    \includegraphics[width=0.48\textwidth]{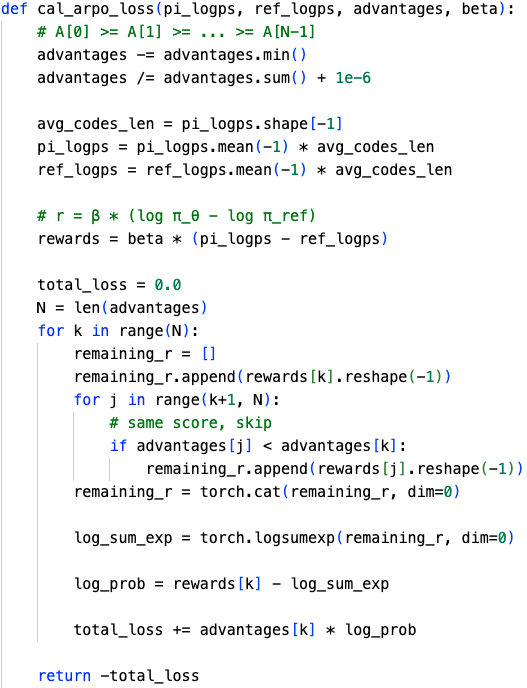}
    \caption{ 
    Python Code of ARPO.
    }
    \label{fig:arpo_code}
\end{figure}
The code of ARPO is given in \cref{fig:arpo_code}.

\subsection{Pseudo-code of Asynchronous ARPO}
The pseudo-code of asynchronous online ARPO with truncated training is given in \cref{alg:async_rl_framework}.

\RestyleAlgo{ruled}
\begin{algorithm}[h]
\caption{Asynchronous Online ARPO with Truncated Training}\label{alg:async_rl_framework}
\SetKwComment{Comment}{// }{}
\SetKwIF{If}{ElseIf}{Else}{if}{:}{elif}{else:}{}%
\SetKwFor{For}{for}{\string:}{}%
\SetKwFor{While}{while}{\string:}{}%
\SetKwProg{Fn}{Procedure}{:}{}
\SetKwFunction{SyncWeights}{SyncWeights}
\SetKwFunction{Truncate}{RandomTruncate}
\SetKwFunction{CalcLoss}{CalcARPOLoss}
\SetKwFunction{Optimize}{OptimizerStep}
\SetKwFunction{Save}{SaveCheckpoint}
\SetKwFunction{Discard}{DiscardOutdatedData}

\KwData{Pretrained Policy $\pi_{\theta}$, Replay Buffer $\mathcal{B}$, Pre-start samples $S_1$, Steps $N_1, N_2$}
\KwResult{Optimized Policy $\pi^*$}
\BlankLine

\Fn{\textbf{RolloutWorker}}{
    \While{\text{True}}{
        $\pi_{\text{local}} \leftarrow \SyncWeights(\pi_{\text{global}})$ \Comment*[r]{Fetch latest policy}
        $M_{\text{full}} \leftarrow \pi_{\text{local}}(P)$ \Comment*[r]{Generate complete mesh}
        $\{M_t\}_{t=1}^{\mathcal{T}} \leftarrow \Truncate(M_{\text{full}})$ \Comment*[r]{Truncated Training}
        \For{each $M_t \in \{M_t\}$}{
            $R_t \leftarrow \text{ComputeReward}(M_t)$\;
            $\mathcal{B}.\text{push}(M_t, P, R_t, \text{version}=V)$\;
        }
    }
}
\BlankLine

\Fn{\textbf{TrainerWorker}}{
    $V \leftarrow 0$\;
    \Comment{Pre-Start Stage for Stable Training}
    \While{$|\mathcal{B}| < S_1$}{
        Wait for initial samples\;
    }
    \For{$step \leftarrow 1$ \KwTo $N_1$}{
        Batch $b \leftarrow \mathcal{B}.\text{sample()}$\;
        $\mathcal{L}_{\text{ARPO}} \leftarrow \CalcLoss(\pi_{\text{global}}, b)$\;
        $\pi_{\text{global}} \leftarrow \Optimize(\pi_{\text{global}}, \mathcal{L}_{\text{ARPO}})$\;
    }
    $\Save(\pi_{\text{global}}, V+1)$\;
    $\Discard(\mathcal{B}, \text{version} \le 0)$ \Comment*[r]{Discard pre-start data}
    $V \leftarrow V + 1$\;
    
    \BlankLine
    \Comment{Asynchronous Online Loop}
    \While{\text{True}}{
        \For{$step \leftarrow 1$ \KwTo $N_2$}{
            \Comment{Sample data from Policy $V-1$}
            Batch $b \leftarrow \mathcal{B}.\text{sample}(\text{target\_ver}=V-1)$\;
            $\mathcal{L}_{\text{ARPO}} \leftarrow \CalcLoss(\pi_{\text{global}}, b)$\;
            $\pi_{\text{global}} \leftarrow \Optimize(\pi_{\text{global}}, \mathcal{L}_{\text{ARPO}})$\;
        }
        $\Save(\pi_{\text{global}}, V+1)$\;
        $\Discard(\mathcal{B}, \text{keep\_ver}=V)$ \Comment*[r]{Maintain consistency}
        $V \leftarrow V + 1$\;
    }
}

\end{algorithm}

\subsection{Asynchronous ARPO Subsumes Asynchronous DPO as A Special Case}
For a truncated mesh $M_t$, the truncated asynchronous ARPO objective is represented as 
\begin{equation}
\mathcal{L}_{\mathrm{t-ARPO}}
= -\,\mathbb{E}_{M_t}\Bigg[\sum_{i=1}^K A_t^{(i)}
\log\frac{\exp\!\Big(\mathcal{R}_{i|m:m+w}\Big)}
{\sum_{j=i}^K \exp\!\Big(\mathcal{R}_{j|m:m+w}\Big)}\Bigg], 
\end{equation} 
\begin{equation}
\mathcal{R}_{i|m:m+w} = \beta \log\frac{\pi_\theta(y_{i|m:m+w}\mid P)}{\pi_{\mathrm{ref}}(y_{i|m:m+w}\mid P)},
\end{equation} 
where $m$ and $w$ are the initial position and truncated window length, respectively. $P$ is the input point cloud. Given rewards $\{R^{(i)}\}_{i=1}^K$ (smaller subscript means stronger preference), the corresponding advantages are 
\begin{equation}
A^{(k)} = \frac{R^{(k)} - \min\{\{R^{(i)}\}_{i=1}^K\} }{\sum_{k=1}^K (R^{(k)} - \min\{\{R^{(i)}\}_{i=1}^K\}) + \epsilon}.
\label{eq:advantage}
\end{equation}

When $\text{group}=2$, the advantage function, after equation \cref{eq:advantage} computation, exhibits only two cases: (i) When $R^{(1)} \neq R^{(2)}$, then $A^{(1)}=1$ and $A^{(2)}=0$. (ii) Otherwise, the two advantage values are equal to 0. In case (ii), we exclude this case from policy updates. Meanwhile, when group is set to 2, the Plackett--Luce model \cite{PL_model} degenerates into the Bradley-Terry model \cite{BT_model} 
\begin{align}
\mathcal{L}_{\mathrm{t-ARPO}}
&= -\,\mathbb{E}\Bigg[\sum_{i=1}^2 A_t^{(i)}
\log\frac{\exp\!(\mathcal{R}_{i|m:m+w})}
{\sum_{j=i}^2 \exp\!(\mathcal{R}_{j|m:m+w})}\Bigg], \notag \\
&= -\,\mathbb{E}\Bigg[1 \times \log\frac{\exp\!(\mathcal{R}_{1|m:m+w})} {\sum_{j=1}^2 \exp\!(\mathcal{R}_{j|m:m+w})} \notag \\ &+ 0 \times \log\frac{\exp\!(\mathcal{R}_{2|m:m+w})} {\sum_{j=2}^2 \exp\!(\mathcal{R}_{j|m:m+w})}\Bigg] \notag \\
&= -\,\mathbb{E}\Bigg[\log\frac{\exp\!(\mathcal{R}_{1|m:m+w})} {\exp\!(\mathcal{R}_{1|m:m+w}) + \exp\!(\mathcal{R}_{2|m:m+w})}\Bigg].
\label{eq:de_ARPO}
\end{align}
\cref{eq:de_ARPO} can be organized as 
\begin{align}
\mathcal{L}_{\mathrm{t-ARPO}} 
&= -\,\mathbb{E}_{M_t}\Bigg[\log \sigma (\mathcal{R}_{1|m:m+w} - \mathcal{R}_{2|m:m+w}) \Bigg] \notag \\
& = -\,\mathbb{E}_{M_t}\Bigg[\log \sigma (\mathcal{R}_{\mathcal{W}|m:m+w} - \mathcal{R}_{\mathcal{L}|m:m+w}) \Bigg] \notag \\
&= -\,\mathbb{E}_{M_t}\Bigg[\log \sigma \bigg(\beta \log\frac{\pi_\theta(y_{\mathcal{W}|m:m+w}\mid P)}{\pi_{\mathrm{ref}}(y_{\mathcal{W}|m:m+w}\mid P)} \notag \\
&- \beta \log\frac{\pi_\theta(y_{\mathcal{L}|m:m+w}\mid P)}{\pi_{\mathrm{ref}}(y_{\mathcal{L}|m:m+w}\mid P)} \bigg) \Bigg] \notag \\
&= \mathcal{L}_{\mathrm{t-DPO}}.
\end{align}
where $y_{\mathcal{W}|m:m+w}$ and $y_{\mathcal{L}|m:m+w}$ represent winner and loser, respectively. That is, asynchronous ARPO degenerates into asynchronous DPO. Both ARPO and DPO are fundamentally built upon implicit reward modeling and preference-based policy optimization, leading to fast and stable training convergence. ARPO further introduces explicit advantage function guidance to improve generalization ability and the quality of generated meshes.

\subsection{Reward Design}
The reward function $R(M_t)$ for a generated mesh $M_t$ given a prompt $P$ is 
\begin{align}
\scalebox{0.93}{$
R(M_t) = \begin{cases} w_{\text{qr}} \cdot N_{\text{qr}} + N^2_{\text{ql}} & \text{if } N_\text{bf} < \theta_{ray} \text{ and } D_\text{hd} < \theta_{hd}, \\ 0 & \text{otherwise}. \end{cases}
$}
\label{eq:reward}
\end{align}

\subsubsection{Ray-based Reward}
The core objective of ray-based integrity reward $R_{\text{ray}}(M_t)$ is to identify a set of ``bad faces'', denoted as $\mathcal{F}_{\text{bad}}$, which indicate incomplete or inconsistent regions. The size of this set, $N_\text{bf} = |\mathcal{F}_{\text{bad}}|$, is then used to compute the final reward. The process is outlined in \cref{alg:bad_face_identification}. 

\RestyleAlgo{ruled}
\begin{algorithm}[h]
\caption{Identification of Bad Faces for Ray-based Reward}\label{alg:bad_face_identification}
\SetKwComment{Comment}{// }{}
\SetKwIF{If}{ElseIf}{Else}{if}{:}{elif}{else:}{}%
\SetKwFor{For}{for}{\string:}{}%
\SetKwFunction{RayIntersect}{RayMeshIntersection}
\SetKwFunction{GetBoundaryFaces}{GetTopologicalBoundaryFaces}
\SetKwFunction{GetIncidentFaces}{GetIncidentFaces}

\KwData{Mesh $\mathcal{M}=(\mathcal{V}, \mathcal{F})$, angle threshold $\theta_{\text{angle}}=0$, ratio gate $\theta_{\text{ratio}}=0.0005$}
\KwResult{Set of bad faces $\mathcal{F}_{\text{bad}}$}
\BlankLine

\Comment{Global Integrity Pre-Check}
$\mathcal{R} \leftarrow \text{GenerateOrthogonalRays()}$\;
$\text{hits} \leftarrow \RayIntersect(\mathcal{M}, \mathcal{R})$\;
$n_{\text{invalid}} \leftarrow \sum_{(\mathbf{n}, \mathbf{d}) \in \text{hits}} \mathbb{I}(\mathbf{n} \cdot (-\mathbf{d}) < \theta_{\text{angle}})$ \Comment*[r]{Count invalid (back-face) hits}
\If{$n_{\text{invalid}} / |\text{hits}| \le \theta_{\text{ratio}}$}{
    \KwRet{$\emptyset$} \Comment*[r]{Mesh passes pre-check; $N_\text{bf}=0$}
}
\BlankLine

\Comment{Fine-Grained Bad Vertex Localization}
$\mathcal{V}_{\text{bad}} \leftarrow \emptyset$\;
$\mathcal{P}_{\text{view}} \leftarrow \text{DefineExternalViewpoints()}$\;
\For{each vertex $\mathbf{v}_i \in \mathcal{V}$}{
    is\_vertex\_bad $\leftarrow$ \textbf{false}\;
    \For{each viewpoint $\mathbf{p}_j \in \mathcal{P}_{\text{view}}$}{
        \If{$\text{IsVisible}(\mathcal{M}, \mathbf{v}_i, \mathbf{p}_j)$}{
            \Comment{Probe the vertex's local neighborhood}
            $\mathcal{T} \leftarrow \text{GeneratePointsInNeighborhood}(\mathbf{v}_i)$\;
            $\mathcal{R}' \leftarrow \text{RaysFromViewpointToTargets}(\mathbf{p}_j, \mathcal{T})$\;
            $\text{probe\_hits} \leftarrow \RayIntersect(\mathcal{M}, \mathcal{R}')$\;
            $n'_{\text{invalid}} \leftarrow \sum_{(\mathbf{n}', \mathbf{d}') \in \text{probe\_hits}} \mathbb{I}(\mathbf{n}' \cdot (-\mathbf{d}') < \theta_{\text{angle}})$\;
            \If{$n'_{\text{invalid}} > 0$}{
                is\_vertex\_bad $\leftarrow$ \textbf{true}\;
                \textbf{break} \Comment*[r]{Break from viewpoints loop}
            }
        }
    }
    \If{is\_vertex\_bad}{
        $\mathcal{V}_{\text{bad}} \leftarrow \mathcal{V}_{\text{bad}} \cup \{\mathbf{v}_i\}$\;
    }
}
\BlankLine

\Comment{Bad Face Set Identification}
$\mathcal{F}_{\text{candidate}} \leftarrow \GetIncidentFaces(\mathcal{V}_{\text{bad}})$\;
$\mathcal{F}_{\text{boundary}} \leftarrow \GetBoundaryFaces(\mathcal{M})$ \Comment*[r]{Faces incident to open edges}
$\mathcal{F}_{\text{bad}} \leftarrow \mathcal{F}_{\text{candidate}} \cap \mathcal{F}_{\text{boundary}}$\;
\KwRet{$\mathcal{F}_{\text{bad}}$}\;

\end{algorithm}

\noindent\textbf{Global Integrity Pre-Check.}
First, we perform a coarse-grained test to quickly assess the mesh's overall integrity. We cast a dense grid of parallel rays towards the mesh from multiple orthogonal directions. For each intersection, we compare the surface normal $\mathbf{n}$ with the incoming ray direction $\mathbf{d}$. An intersection is deemed invalid if the normal points away from the ray's origin ($\mathbf{n} \cdot (-\mathbf{d}) < \theta_{\text{angle}}$), indicating a potential back-face hit. If the ratio of these invalid hits is below a gate threshold $\theta_{\text{ratio}}$, the mesh is considered intact ($N_\text{bf} = 0$), and the expensive bad face localization stage is skipped.

\begin{figure}[t]
    \centering
    \includegraphics[width=0.48\textwidth]{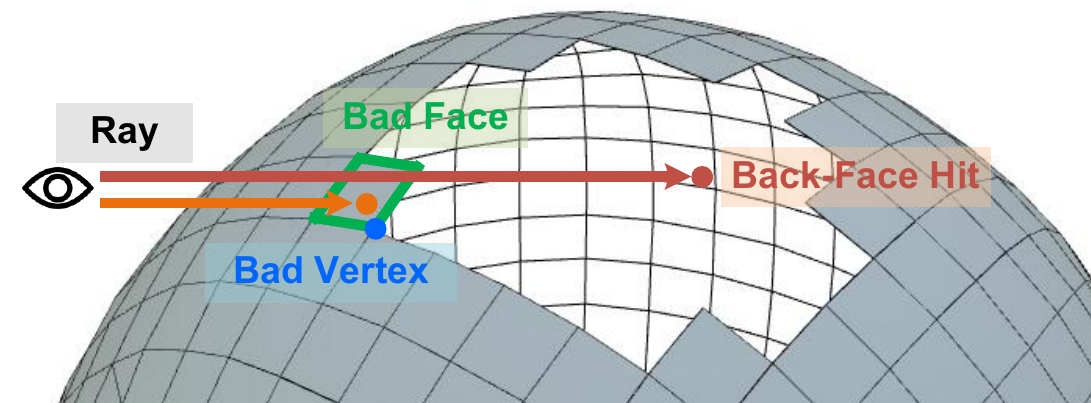}
    \caption{ 
    Illustration of ray casting integrity check. 
    When a ray is cast toward the mesh from a given direction, it may pass through the broken region surrounding a ``bad face'', leading to a direct back-face hit on the mesh. The number of bad faces $N_\text{bf}$ reflects the extent of geometric breakage in the generated mesh.
    }
    \label{fig:ray}
\end{figure}

\noindent\textbf{Fine-Grained Bad Vertex Localization.}
If the mesh fails the pre-check, we proceed to pinpoint the specific vertices contributing to the inconsistency and incompleteness. We first establish a set of fixed external viewpoints and identify all vertices visible from these locations. For each visible vertex, we probe its local geometry using a perturbation strategy: instead of casting one ray, we cast numerous rays from the viewpoint to random target points within a small spherical neighborhood around the vertex, as shown in \cref{fig:ray}. This enhances robustness against singularities. A vertex is labeled as ``bad'' if any of these probing rays result in an invalid back-facing hit, suggesting it lies on a geometric discontinuity.

\noindent\textbf{Bad Face Set Identification.}
Finally, the set of identified bad vertices ($\mathcal{V}_{\text{bad}}$) is used to form a candidate set of bad faces ($\mathcal{F}_{\text{candidate}}$). To refine this set and ensure geometric inconsistencies align with topological breaks, we apply a filter. We compute the set of all true boundary faces ($\mathcal{F}_{\text{boundary}}$) by identifying edges incident to only a single face. The final set of bad faces, $\mathcal{F}_{\text{bad}}$, is the intersection of the ray-detected candidates and the topologically-identified boundary. The size of this set, $N_\text{bf} = |\mathcal{F}_{\text{bad}}|$, is the value used to determine the reward $R_{\text{ray}}(M_t)$ by comparing it against the threshold $\theta_{ray}$. If $N_\text{bf}$ exceeds $\theta_{ray}$, it means that the current mesh is incomplete and has broken faces.

\noindent\textbf{Geometric Integrity Reward for Truncated Mesh.}
At this point, we have calculated the location indices of all bad faces in the mesh before truncation. During truncation training, we directly use the bad faces within the truncation region to calculate the ray-based reward $R_{\text{t-ray}}(M_t)$.

\begin{figure}[t]
    \centering
    \includegraphics[width=0.48\textwidth]{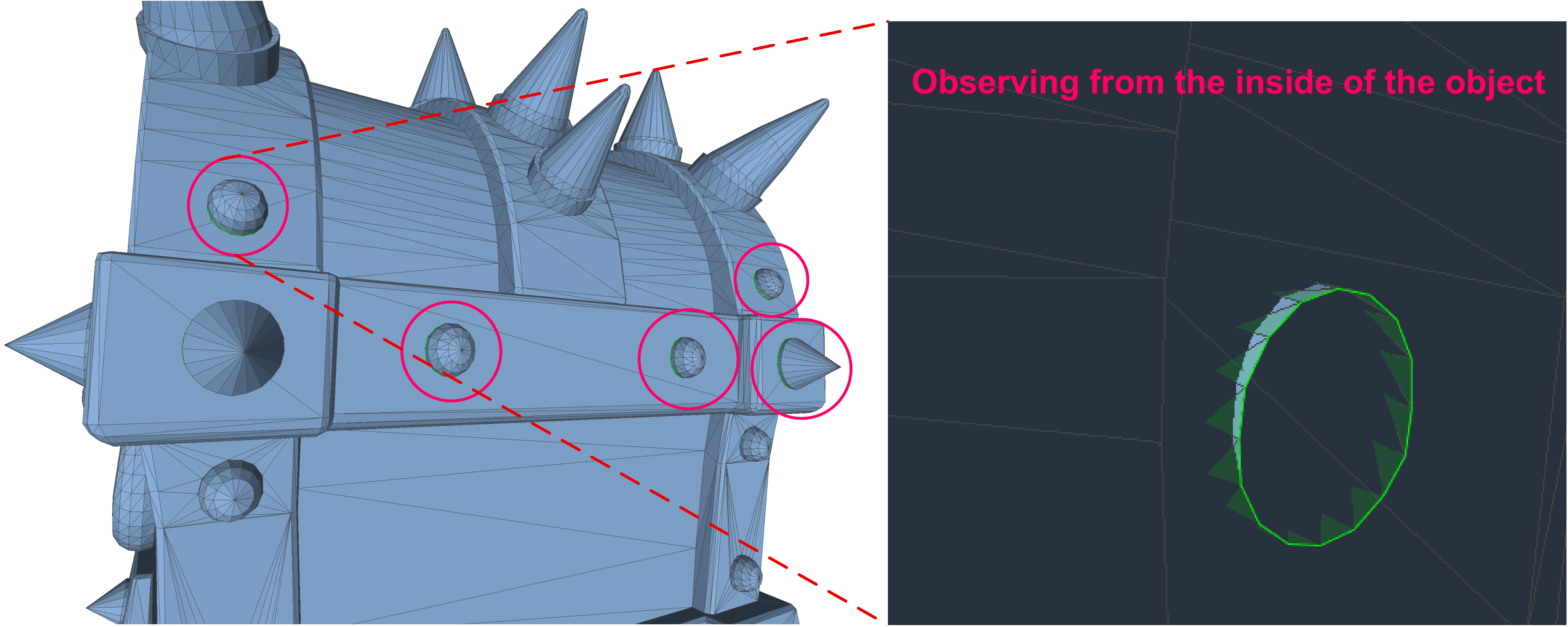}
    \caption{ 
    Boundary edge–based reward leads to misjudgments of the multi-component object.
    When a mesh consists of multiple components, boundary edges (highlighted by the green line) often appear between components. However, this mesh is still a good output and should be encouraged. In Mesh-Pro, the ray-based reward does not suffer from this issue.
    }
    \label{fig:ray_vs_be}
\end{figure}

\noindent\textbf{Comparison with Boundary Edge-based Reward.}
Previous methods \cite{liu2025meshrft, liu2025quadgpt} primarily rely on boundary edges to detect fractures and formulate corresponding reward functions. However, such boundary edge-based reward is prone to misjudgment when dealing with multi-component objects. As illustrated in \cref{fig:ray_vs_be}, for a good multi-component mesh, the model should ideally be incentivized to increase its output probability, yet the frequent occurrence of boundary edges between components often results in erroneously low rewards. In Mesh-Pro, the ray-based reward mechanism effectively avoids this issue.

\begin{figure}[t]
    \centering
    \includegraphics[width=0.25\textwidth]{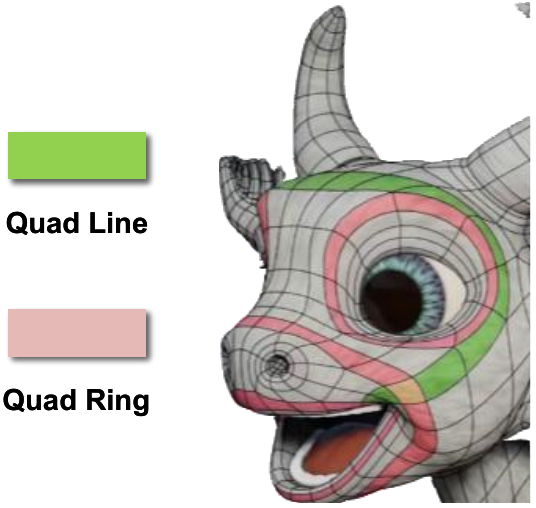}
    \caption{ 
    Illustration of quad rings and quad lines. 
    }
    \label{fig:topo}
\end{figure}

\subsubsection{Topological Reward}
Drawing inspiration from \cite{liu2025quadgpt}, we posit that a well-structured edge flow is a primary characteristic of a high-quality quadrilateral topology. This quality is assessed by quantifying two fundamental topological structures: quad rings and quad lines. \cref{alg:topo_reward} provides the pseudocode for the topological reward $R_{\text{topo}}(M_t)$. Quad rings are identified as closed sequences of quadrilateral faces, whereas quad lines represent open-ended strips of such faces. To distinguish between these, we implement an edge-based traversal algorithm that navigates across adjacent quadrilaterals. Traversal paths that conclude upon reaching a boundary are categorized as quad lines, while paths that loop back to their originating edge are defined as quad rings. The topological reward, denoted as $R_{\text{topo}}(M_t)$, is then formulated as a weighted sum of the total count of quad rings ($N_{\text{qr}}$) and the squared count of quad lines ($N^2_{\text{ql}}$) within the evaluated truncated mesh region (see \cref{eq:reward}). \cref{fig:topo} provides a diagram of quad rings and quad lines.

\RestyleAlgo{ruled}
\begin{algorithm}[t]
\caption{Topological Reward Calculation}\label{alg:topo_reward}
\SetKwComment{Comment}{// }{}
\SetKwIF{If}{ElseIf}{Else}{if}{:}{elif}{else:}{}%
\SetKwFor{For}{for}{\string:}{}%
\SetKwFunction{GetAdjQuad}{GetAdjacentQuad}
\SetKwFunction{GetOppEdge}{GetOppositeEdge}

\KwData{Truncated Mesh $\mathcal{M}_t$, ring weight $w_{\text{qr}}$}
\KwResult{Topological reward $R_{\text{topo}}$}
\BlankLine

\Comment{Initialize counters and set of visited edges}
$N_{\text{qr}} \leftarrow 0$; $N_{\text{ql}} \leftarrow 0$\;
$\mathcal{E}_{\text{processed}} \leftarrow \emptyset$\;
\BlankLine

\For{each edge $\mathbf{e}_{\text{start}} \in \mathcal{M}_t$}{
    \If{$\mathbf{e}_{\text{start}} \notin \mathcal{E}_{\text{processed}}$}{
        \Comment{Begin traversal from an unprocessed edge}
        path\_edges $\leftarrow \emptyset$\;
        $\mathbf{e}_{\text{current}} \leftarrow \mathbf{e}_{\text{start}}$\;

        \While{$\mathbf{e}_{\text{current}} \neq \text{null}$ \textbf{and} $\mathbf{e}_{\text{current}} \notin \text{path\_edges}$}{
            path\_edges.add($\mathbf{e}_{\text{current}}$)\;
            $f_{\text{adj}} \leftarrow \GetAdjQuad(\mathbf{e}_{\text{current}})$ \Comment*[r]{Find adjacent quad face}
            \If{$f_{\text{adj}}$ exists}{
                $\mathbf{e}_{\text{current}} \leftarrow \GetOppEdge(\mathbf{e}_{\text{current}}, f_{\text{adj}})$\;
            } \Else{
                $\mathbf{e}_{\text{current}} \leftarrow \text{null}$ \Comment*[r]{Path terminates at boundary}
            }
        }
        
        \Comment{Mark all edges in the path as processed}
        $\mathcal{E}_{\text{processed}} \leftarrow \mathcal{E}_{\text{processed}} \cup \text{path\_edges}$\;

        \Comment{Classify path as a ring or a line}
        \If{$\mathbf{e}_{\text{current}} == \mathbf{e}_{\text{start}}$}{
            $N_{\text{qr}} \leftarrow N_{\text{qr}} + 1$\;
        } \Else{
            $N_{\text{ql}} \leftarrow N_{\text{ql}} + 1$\;
        }
    }
}
\BlankLine

\Comment{Return the final weighted topological reward}
$R_{\text{topo}}=w_{\text{qr}} \cdot N_{\text{qr}} + N_{\text{ql}}^2$

\KwRet{$R_{\text{topo}}$}\;
\end{algorithm}

\section{Experiments}
\subsection{Dataset Construction}
\noindent\textbf{Pre-training Dataset Construction.}
A large-scale, high-fidelity dataset is foundational to the pre-training strategy, for which we develop a comprehensive data generation pipeline. We begin by aggregating 3D assets from diverse public and professional repositories, including Objaverse~\citep{deitke2023objaverse}, Objaverse-XL~\citep{deitke2023objaversexl}, ShapeNetV2~\citep{chang2015shapenet}, and 3D-FUTURE~\citep{fu20213d}. To significantly expand the availability of quad-dominant meshes, we implement an automated triangle-to-quadrilateral conversion operator. This process intelligently merges triangles into quadrilaterals, followed by a strict structural integrity check step. The resulting augmented dataset is then subjected to a rigorous filtering process. First, a series of rule-based operators (e.g., fractured geometry detection, small face detection) are used to filter out meshes with unreasonable geometric structures. Then, an AI model-based quality assessment operator is trained to score topological quality, which is used to screen for meshes with low-quality topology. Finally, we select models with face counts between 500 and 20,000, assembling our final pre-training dataset of 1.3 million high-quality 3D models.

\noindent\textbf{Reinforcement Learning Dataset Construction.}
We meticulously select a dataset of 500 dense meshes generated by Hunyuan3D 2.5 \cite{lai2025hunyuan3d}, supplemented with 200 artist meshes. This collection features a high degree of both diversity and quality, enabling the asynchronous ARPO phase to comprehensively learn the reward distribution from the geometric and topological structures of these models. During the asynchronous online ARPO training process, point clouds are repeatedly and randomly sampled from these models to generate multiple outputs. This process guides Mesh-Pro to favor the generation of meshes that are associated with higher rewards.

\RestyleAlgo{ruled}
\begin{algorithm}[t]
\caption{Broken Mesh Detection}\label{alg:broken_ratio}
\SetKwComment{Comment}{// }{}
\SetKwIF{If}{ElseIf}{Else}{if}{:}{elif}{else:}{}%
\SetKwFor{For}{for}{\string:}{}%
\SetKwFunction{Normalize}{Normalize}
\SetKwFunction{Raycast}{Raycast}
\SetKwFunction{GetNormal}{GetFaceNormal}

\KwData{Mesh $\mathcal{M}$, angle threshold $\theta_{\text{angle}}$, success threshold $\theta_{\text{succ}}$, random scale $\sigma_{\text{rand}}$}
\KwResult{A boolean value, is\_broken}
\BlankLine

$\mathcal{M}' \leftarrow \Normalize(\mathcal{M})$ \Comment*[r]{Scale mesh to fit in a unit box}
$\mathcal{R}_{\text{origins}} \leftarrow \emptyset$; $\mathcal{R}_{\text{dirs}} \leftarrow \emptyset$\;
\BlankLine

\Comment{Generate rays from 6 directions (±X, ±Y, ±Z)}
\For{each axis $\mathbf{a} \in \{(1,0,0), (0,1,0), (0,0,1)\}$}{
    \For{each sign $s \in \{+1, -1\}$}{
        $\mathcal{O} \leftarrow$ Generate a grid of points on a plane far from $\mathcal{M}'$ along $-s \cdot \mathbf{a}$\;
        $\mathcal{D}_{\text{aligned}} \leftarrow$ Create corresponding axis-aligned directions $s \cdot \mathbf{a}$\;
        $\mathcal{D}_{\text{random}} \leftarrow$ Create perturbed directions from $\mathcal{D}_{\text{aligned}}$ using $\sigma_{\text{rand}}$\;
        \BlankLine
        \Comment{Add both aligned and randomized rays}
        $\mathcal{R}_{\text{origins}} \leftarrow \mathcal{R}_{\text{origins}} \cup \mathcal{O} \cup \mathcal{O}$\;
        $\mathcal{R}_{\text{dirs}} \leftarrow \mathcal{R}_{\text{dirs}} \cup \mathcal{D}_{\text{aligned}} \cup \mathcal{D}_{\text{random}}$\;
    }
}
\BlankLine
\Comment{Perform ray-mesh intersection}
$(\mathcal{I}_{\text{tri}}, \mathcal{I}_{\text{ray}}, \mathcal{L}) \leftarrow \Raycast(\mathcal{M}', \mathcal{R}_{\text{origins}}, \mathcal{R}_{\text{dirs}})$ \Comment*[r]{Find first hits}

\If{$\mathcal{L}$ is empty}{
    \KwRet{False} \Comment*[r]{No intersections found}
}
\BlankLine

$N_{\text{errors}} \leftarrow 0$\;
\For{each hit $i$ from $1$ to $|\mathcal{L}|$}{
    $\mathbf{n} \leftarrow \GetNormal(\mathcal{M}', \mathcal{I}_{\text{tri}}[i])$ \Comment*[r]{Normal of the hit face}
    $\mathbf{d} \leftarrow \mathcal{R}_{\text{dirs}}[\mathcal{I}_{\text{ray}}[i]]$ \Comment*[r]{Direction of the hitting ray}
    
    \If{$\mathbf{n} \cdot (-\mathbf{d}) < \theta_{\text{angle}}$}{
        $N_{\text{errors}} \leftarrow N_{\text{errors}} + 1$\;
    }
}
\BlankLine

\Comment{Classify mesh based on the error ratio}
score $\leftarrow N_{\text{errors}} / |\mathcal{L}|$\;
is\_broken $\leftarrow$ score $> \theta_{\text{succ}}$\;

\KwRet{is\_broken}\;
\end{algorithm}

\subsection{Evaluation Metrics}
\subsubsection{Broken Ratio Evaluation Criteria}
We propose a robust method for detecting geometric defects in a generated mesh, as described in \cref{alg:broken_ratio}. The core of our approach is based on ray casting and analyzing the orientation of surface normals at intersection points. First, the input mesh is normalized to fit within a unit bounding box. This step ensures a consistent scale for the subsequent ray-casting process. We then generate a dense grid of parallel rays originating from planes positioned outside the bounding box along the three principal axes (X, Y, and Z), in both positive and negative directions. To enhance robustness against glancing hits or specific mesh alignments, we augment these axis-aligned rays with a second set of rays whose directions are slightly perturbed by adding a small random vector.

These rays are then cast towards the mesh, and we compute the first intersection point for each one. For each successful intersection, we evaluate the consistency of the surface normal. For a well-formed, closed manifold, a ray entering the surface should intersect a face whose normal points away from the ray's direction (i.e., outwards from the mesh). This is verified by computing the dot product between the face normal $\mathbf{n}$ and the negative ray direction $-\mathbf{d}$. If this value is below a specified angle threshold $\theta_{\text{angle}}=0$, it suggests an inverted face or a similar topological inconsistency, and the hit is flagged as a ``normal error''.

We compute a fracture score as the ratio of the total number of normal errors to the total number of ray-mesh intersections. If this score exceeds a success threshold $\theta_{\text{succ}}=0.01$, the mesh is classified as topologically broken. The Broken Ratio (BR) is computed as the average of the fracture score (1 for fractured, 0 otherwise) across all generated meshes.

\subsubsection{User Study Evaluation Criteria}
For relative preference scoring, technical art experts are asked to sort all generated meshes from highest to lowest. The main considerations for the User Study (US) are the broken ratio and topological quality. Topological quality mainly assesses the structural regularity of the geometry and edge-flow patterns. In cases with few broken areas, the primary comparison is topological quality. If there are too many broken areas, the mesh quality is considered poor. Human experts rank the generated meshes of $U$ methods from best ($U-1$) to worst ($0$), revealing relative preference relationships.

For absolute quality scoring, technical art experts are asked to assign an integer score between 1 and 5 (higher is better) based on the broken ratio and topological quality of each generated mesh.

Since there are no well-established objective quantitative metrics for judging the artist-like topology and edge flow of generated meshes, we use US as an important reference for the evaluation.

\begin{figure}[t]
    \centering
    \includegraphics[width=0.48\textwidth]{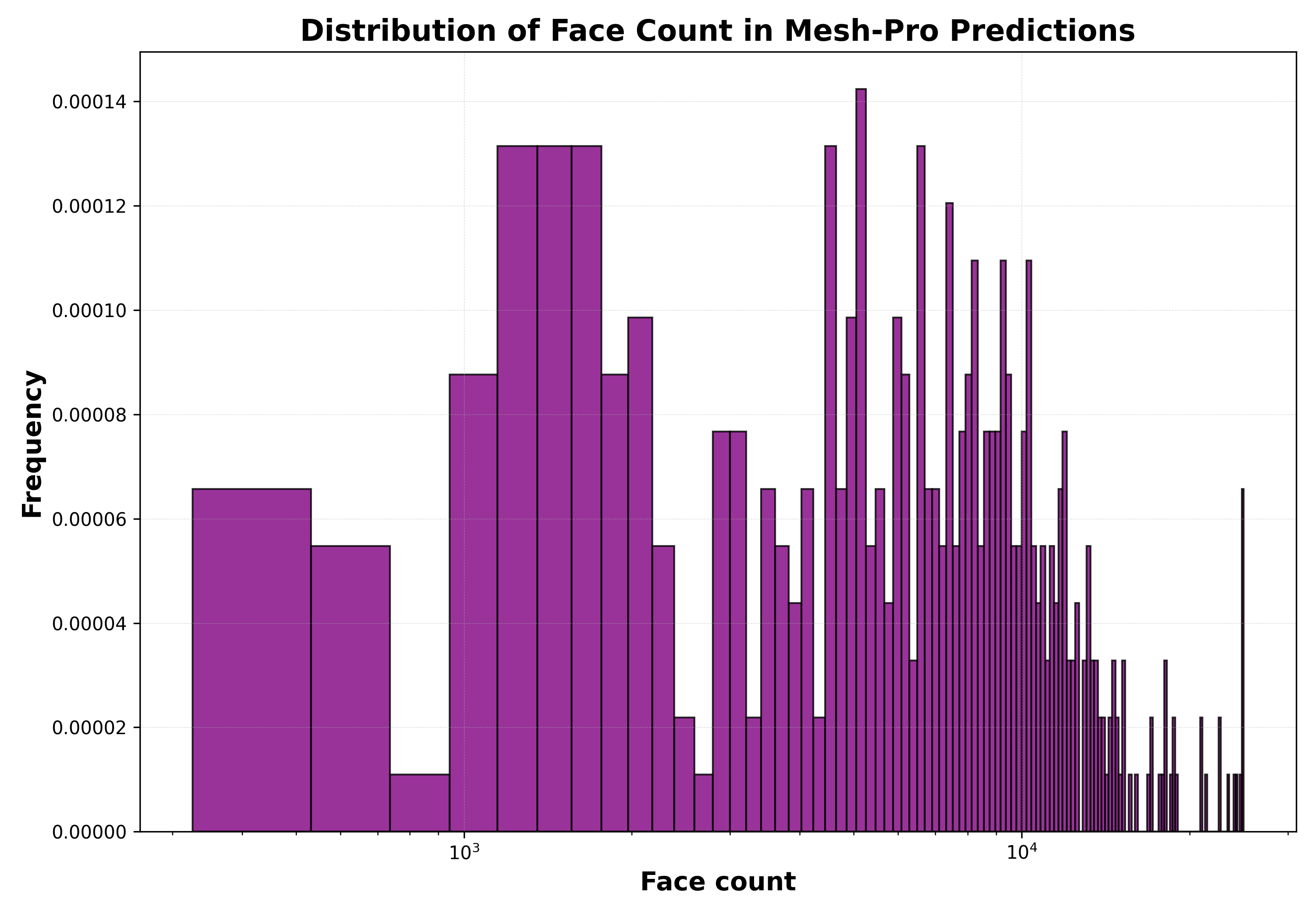}
    \caption{ 
    Distribution of face count (consisting of a mixture of triangles and quadrilaterals) in Mesh-Pro predictions. Point clouds are sampled from dense meshes and artist meshes. The average face count is approximately 8k.
    }
    \label{fig:face_count}
\end{figure}

\begin{figure}[t]
    \centering
    \includegraphics[width=0.48\textwidth]{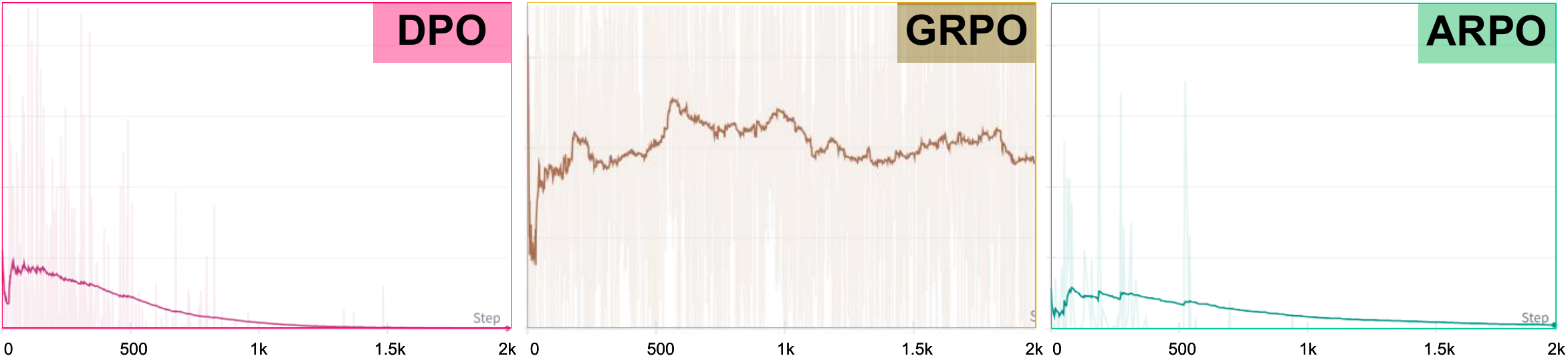}
    \caption{ 
    Training loss curves for DPO, GRPO, and ARPO.
    }
    \label{fig:loss}
\end{figure}

\begin{table}[h]
\centering
\setlength{\tabcolsep}{3pt}
\caption{Broken ratio sensitivity analysis.}
\resizebox{0.54\columnwidth}{!}{ 
    \begin{tabular}{@{}l|ccc@{}}
    \toprule  
    $\theta_{\text{succ}}$  & 0.01 & 0.05 & 0.1 \\
    \midrule  
    MeshAnyv2      & 94\% & 78\% & 62\% \\
    DeepMesh       & 91\% & 81\% & 69\% \\
    FastMesh       & 99\% & 98\% & 95\% \\
    Mesh-Pro       & \textbf{22\%} & \textbf{7\%} & \textbf{1\%} \\
    \bottomrule  
    \end{tabular}
}
\label{tab:BR_sensitivity_analysis}
\end{table}

\begin{table}[t]
\centering
\setlength{\tabcolsep}{3pt}
\caption{Runtime Analysis of Mesh-Pro.}
\resizebox{0.85\columnwidth}{!}{
    \begin{tabular}{@{}lccc@{}}
    \toprule  
    Configuration & Baseline & +KV caching & + CUDA graph \\ 
    \midrule  
    Token/s $\uparrow$ & 40 & 110 & 310 \\
    \bottomrule  
    \end{tabular}
}
\label{tab:Runtime}
\end{table}

\begin{table}
\centering
\setlength{\tabcolsep}{3pt}
\caption{More quantitative comparison on Dense and Artist Meshes.}
\resizebox{0.9\columnwidth}{!}{
    \begin{tabular}{@{}l|ccc|ccc@{}}
    \toprule  
    Data Type &\multicolumn{3}{c|}{Dense Meshes} & \multicolumn{3}{c@{}}{Artist Meshes} \\
    \midrule  
    Metrics & CD $\downarrow$ & NC $\uparrow$ & F1 $\uparrow$ & CD $\downarrow$ & NC $\uparrow$ & F1 $\uparrow$ \\
    \midrule  
    MeshAnyv2~\citep{chen2024meshanythingv2}    & 0.148 & 0.679 & 0.064 & 0.084 & 0.785 & 0.134 \\
    BPT~\citep{weng2025scaling}                 & 0.109 & 0.737 & 0.151 & 0.046 & 0.876 & 0.311 \\
    DeepMesh~\citep{zhao2025deepmesh}           & 0.351 & 0.535 & 0.027 & 0.336 & 0.579 & 0.082 \\
    FastMesh~\citep{kim2025fastmesh}            & 0.099 & 0.714 & 0.112 & 0.044 & 0.849 & 0.254 \\
    Mesh-RFT~\citep{liu2025meshrft}             & 0.051 & 0.809 & 0.208 & 0.041 & 0.885 & 0.328 \\
    QuadGPT~\citep{liu2025quadgpt}              & 0.059 & 0.840 & 0.285 & 0.042 & 0.892 & 0.374 \\
    Mesh-Pro         & \textbf{0.028} & \textbf{0.867} & \textbf{0.316} & \textbf{0.038} & \textbf{0.905} & \textbf{0.391} \\
    \bottomrule  
    \end{tabular}
}
\label{tab:more_quantitative_comparison}
\end{table}

\subsection{Broken Ratio (BR) Criteria Validity}
Ray casting is robust and insensitive (scale-invariant via unit-box and dense omnidirectional casting). In Fig.5 in the main paper and \cref{tab:BR_sensitivity_analysis}, the high BRs (\textgreater90\%) of baselines align with severe visual artifacts (large holes/fragments), confirming they are not false positives but true geometric degradations (watertight meshes achieve $\approx$ 0\% BR). Mesh-Pro demonstrates superior integrity. The significant BR gap persists across varying $\theta_{\text{succ}}$, confirming result robustness. 

For qualitative descriptions regarding the BR in the main paper (e.g., ``consistently generates topologically sound…'', ``exceptional robustness…''), generating strictly valid quad topology is notoriously challenging. Mesh-Pro's 22\% BR stems from minor geometric defects (easily auto-repairable), fundamentally different from the irreparable structural collapse of baselines (see Fig.5 in the main paper). \cref{tab:BR_sensitivity_analysis} confirms this: changing $\theta_{\text{succ}}$ (0.01$\to$0.1) drops our BR to 1\% (near-zero defects), implying negligible minor-defects. Successfully validated in mass industrial game asset production, Mesh-Pro reliably meets strict standards (minor defects are routinely auto-repaired) where baselines fail. Hence, qualitative descriptions are valid.

\subsection{Generated Mesh Face Count Distribution}
By employing the Hourglass Transformer \cite{hao2024meshtron, nawrot2021hierarchical} architecture and truncated training approach (truncated window length is 36,864 tokens), Mesh-Pro is capable of generating meshes across a broad spectrum of face counts. We use point clouds sampled from dense meshes and artist meshes as input. The distribution of face counts (consisting of a mixture of triangles and quadrilaterals) produced by Mesh-Pro is shown in \cref{fig:face_count}. By modeling native quad faces, Mesh-Pro produces clean, well-structured meshes.

\begin{figure*}[t]
    \centering
    \includegraphics[width=1.0\textwidth]{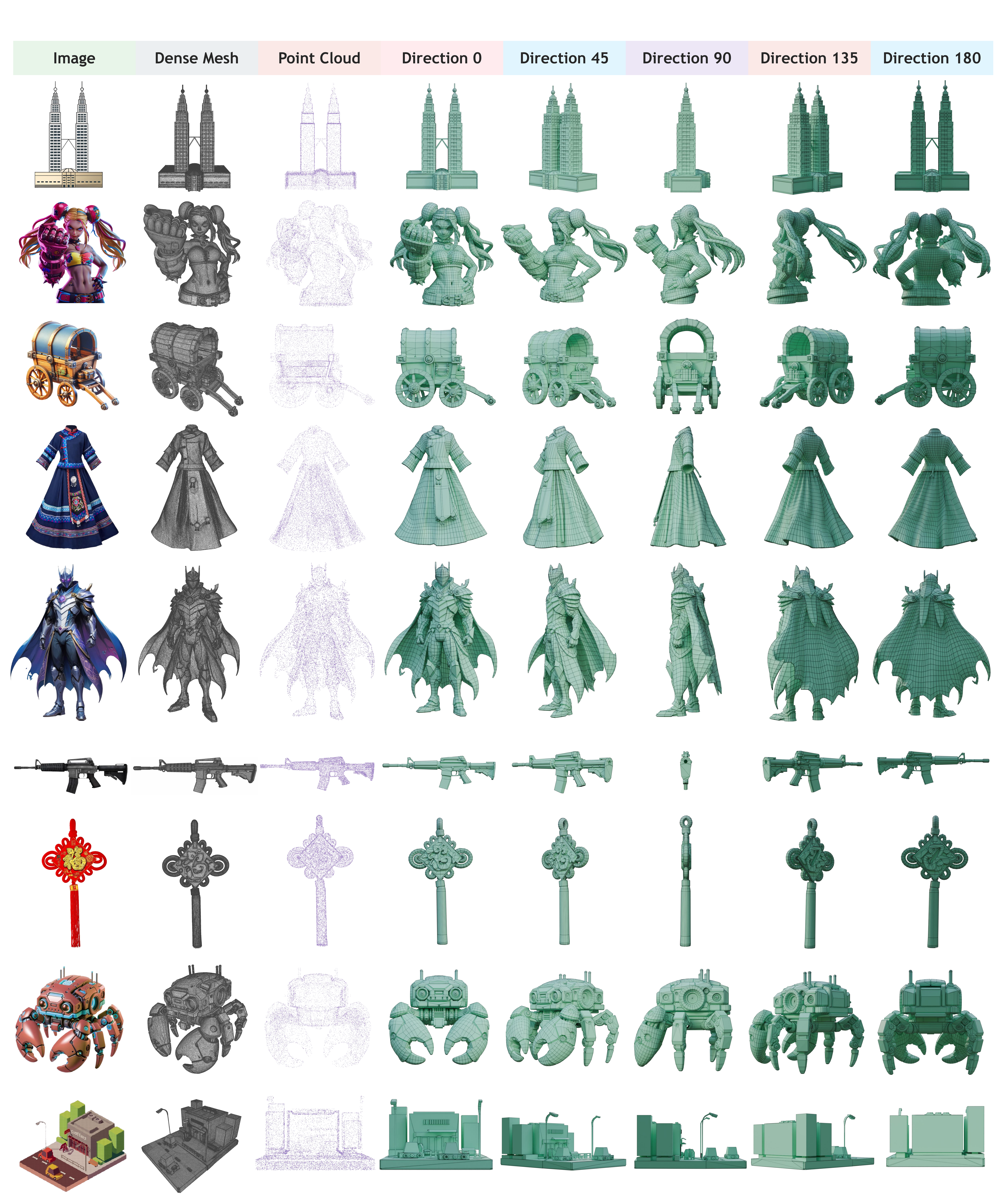}
    \caption{ 
    Multi-view renderings of meshes generated by Mesh-Pro. Mesh-Pro produces diverse meshes with robust geometric integrity and high topological quality.
    }
    \label{fig:image_to_mesh}
\end{figure*}

\begin{figure*}[t]
    \centering
    \includegraphics[width=1.0\textwidth]{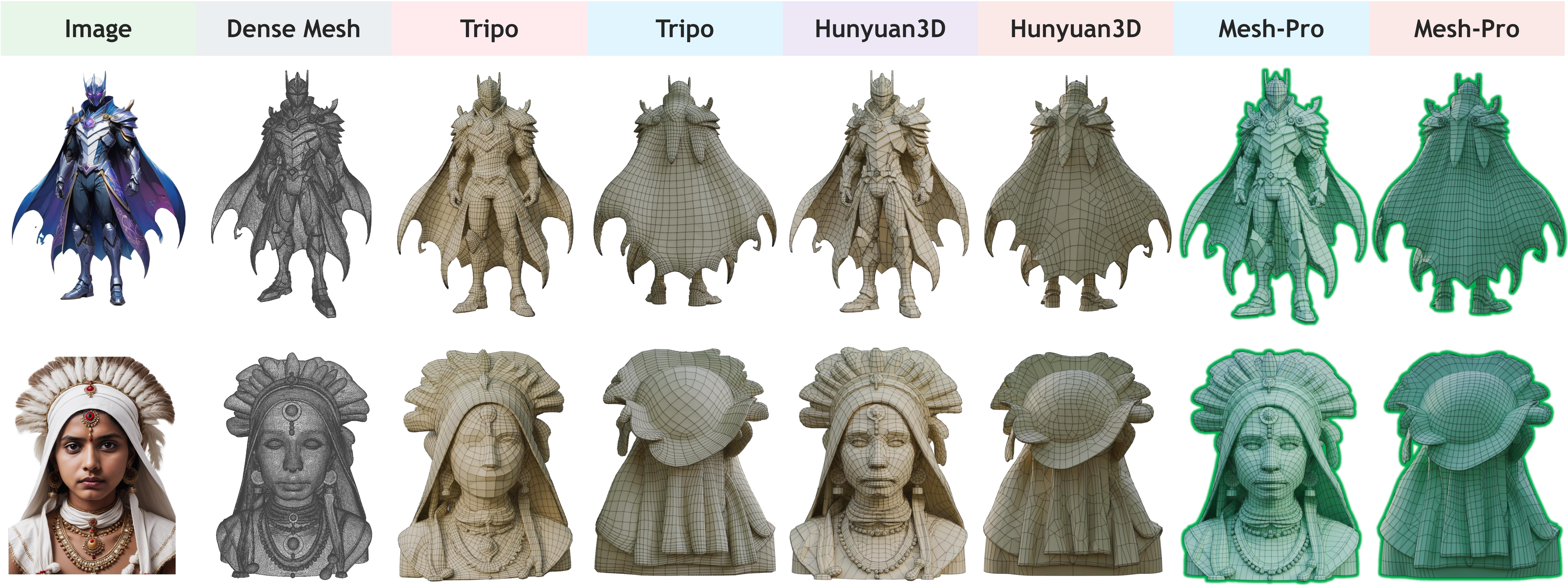}
    \caption{ 
    Qualitative comparison with the closed-source commercial quad mesh generation method Tripo and Hunyuan3D. Mesh-Pro exhibits higher geometric consistency, richer details, and superior topology and edge-flow quality.
    }
    \label{fig:tripo_hunyuan}
\end{figure*}

\begin{figure*}[t]
    \centering
    \includegraphics[width=0.8\textwidth]{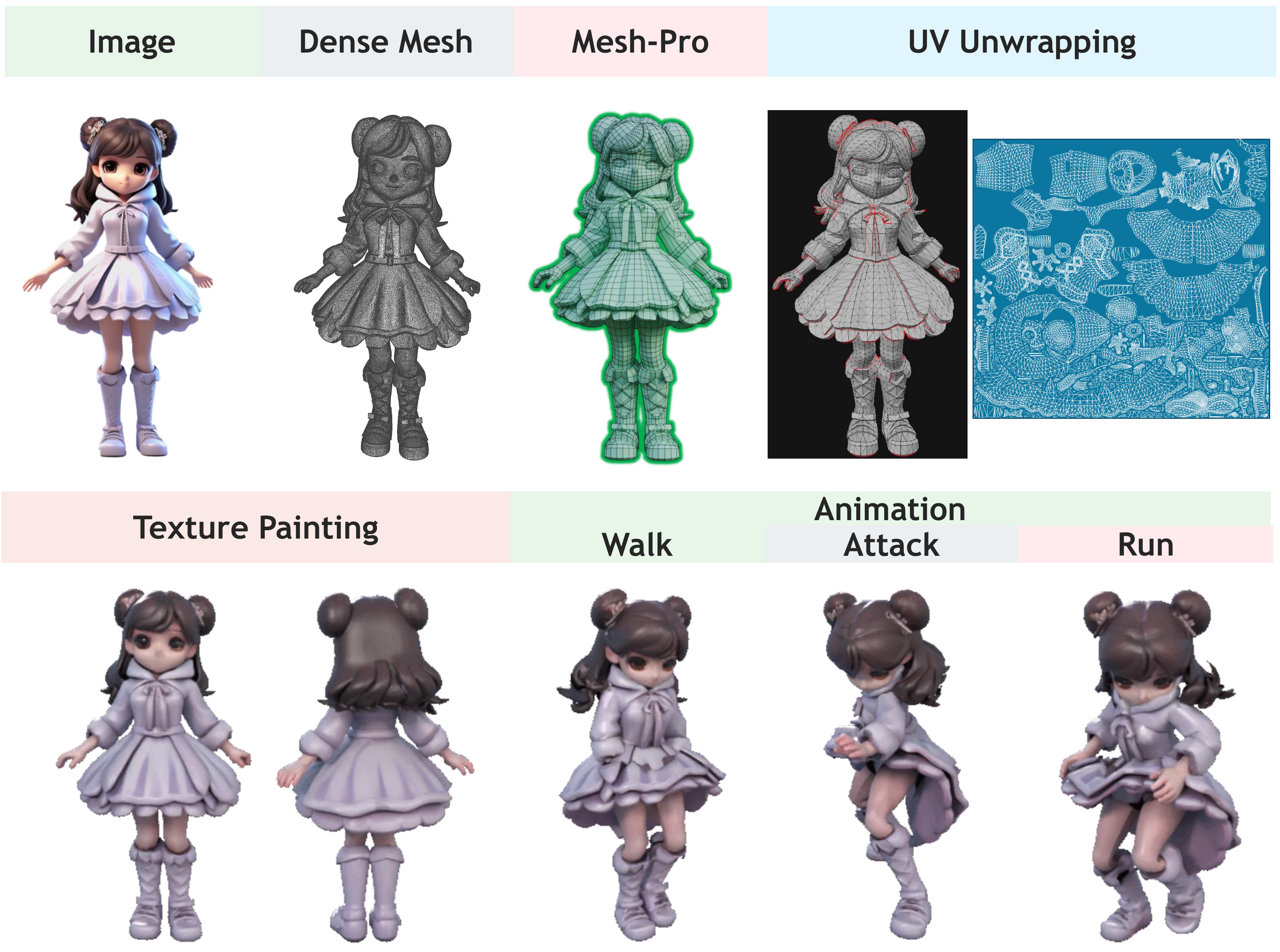}
    \caption{ 
    High-quality native quad-dominant topology generated by Mesh-Pro demonstrates robust performance in downstream tasks, such as UV unwrapping, texture painting, and animation.
    }
    \label{fig:pipeline}
\end{figure*}

\subsection{Loss Curves for DPO, GRPO, and ARPO}
To evaluate the post-training efficiency of different RL algorithms, \cref{fig:loss} plots the training loss curves for DPO, GRPO, and ARPO. Based on ranking preference optimization, both DPO and ARPO achieve stable and fast convergence. In contrast, GRPO converges slowly, as it is constrained by the foundation model's ability to model the complex reward distribution. GRPO suffers from inefficient exploration-exploitation and low training efficiency. Guided by an explicit advantage function, ARPO strikes a superior trade-off between training efficiency and generalization compared to DPO and GRPO, yielding higher-quality mesh generation.

\subsection{Runtime Analysis}
On the NVIDIA H20 GPU, with KV caching and CUDA graph acceleration, the inference speed of Mesh-Pro is approximately 310 token/s, as shown in \cref{tab:Runtime}. According to the proposed quadrilateral tokenization method, 12 tokens contain one triangular face or one quadrilateral face. Based on the calculation method of splitting a quadrilateral face into two triangular faces, and considering that the mesh generated by Mesh-Pro has an average quadrilateral ratio of about 80\%, the inference speed of Mesh-Pro is 47 triangular face/s. Mesh-Pro achieves $\approx$25 quad faces/s, \textless1s detokenization (mesh reconstruction), and E2E latency ($\approx$8k faces) is $\approx$5 min, sufficient for offline industrial asset production.

\subsection{More Quantitative Analysis}
Following existing works \cite{chen2024meshanythingv2}, we report more quantitative comparisons on dense meshes generated by Hunyuan3D 2.5 \cite{lai2025hunyuan3d} and artist meshes selected from the public Toys4k \cite{Toys4k} dataset. As shown in \cref{tab:more_quantitative_comparison}, normal consistency (NC) evaluates the quality of the surface normals. ``CD'' means Chamfer distance. For the F1 score, we build two KD-trees to perform bidirectional nearest-neighbor queries. A point is considered correctly matched if its nearest-neighbor distance is below a threshold (i.e., 0.01). Precision and recall are computed from the predicted-to-GT and GT-to-predicted matches, and F1 is their harmonic mean. Mesh-Pro outperforms existing methods on all metrics, demonstrating superior geometric consistency and more robust mesh generation.

\subsection{More Qualitative Results}
\cref{fig:image_to_mesh} demonstrates multi-view renderings of several samples generated by Mesh-Pro. For image-conditioned generation, we first utilize Hunyuan3D 2.5 \cite{lai2025hunyuan3d} for image-to-3D dense mesh generation (the average face count of dense meshes is 500k). Then, Mesh-Pro samples point clouds from the dense meshes to conduct point-cloud conditioned artist-style quadrilateral mesh generation. Mesh-Pro produces diverse meshes with robust geometric integrity and high topological quality.

Furthermore, we conduct a qualitative comparison with Tripo (a closed-source commercial quad mesh generation method) and Hunyuan3D. As shown in \cref{fig:tripo_hunyuan}, Mesh-Pro exhibits higher geometric consistency, richer details, and superior topology and edge-flow quality.

High-quality native quad-dominant topology generated by Mesh-Pro also demonstrates superior performance in downstream tasks, as illustrated in \cref{fig:pipeline}. This is attributed to its regular topological structure, which facilitates a more efficient UV unwrapping process and a more rational seam layout. Consequently, this allows for high-quality visual effects in texture painting, characterized by less distortion and better consistency. Meanwhile, in animation production, a clear edge flow enables smoother and more controllable mesh deformation, ensuring fluid and natural postures when the model is in motion.

\subsection{Video Demo}
Please refer to the supplementary video for the video demo of Mesh-Pro.

\section{Discussion}
Unlike LLMs (where existing asynchronous RL frameworks are tightly coupled to LLM stacks), mesh token sequences vary more drastically in length, causing severe GPU idle time and training interruptions in synchronous RL. Our asynchronous framework resolves these bottlenecks. Furthermore, mesh generation faces weaker pretrained models and more complex reward distributions (topology vs. geometry). Explicit reward modeling (GRPO) fails to converge and implicit method (DPO) generalizes poorly, ARPO overcomes these issues by leveraging preference optimization for convergence and advantage guidance for generalization. Moreover, asynchronous ARPO is seamlessly applied to other autoregressive models.

\section{Limitations and Future Work}
Despite the significant advantages of Mesh-Pro over existing 3D mesh generation methods, it still has certain limitations and potential avenues for future exploration.

One limitation is that the group size in ARPO is not significantly expanded. This is due to GPU memory constraints, coupled with the limited performance of partial parameter fine-tuning \cite{lora}. We will explore solutions to observe the performance improvement of increasing the group size for asynchronous ARPO in future work. 

Moreover, Mesh-Pro does not yet offer fine-grained control over the face count of generated meshes. Future work will introduce multi-level face-count control into the Truncated Mesh Decoder to better accommodate a wider range of downstream task requirements.

In practice, the rule-based topological edge-flow reward $R_{\text{topo}}(M_t)$ used in Mesh-Pro may lead to reward hacking in the late stage of asynchronous ARPO training, such as producing unnatural edge flow around joint regions. Building on the current rule-based reward, future work will train a reward model with human feedback to further push the upper bound of artist-style quadrilateral mesh generation.

Finally, we hope that the proposed asynchronous ARPO framework will stimulate research on RL algorithms for 3D mesh generation, and that RL in 3D generation can achieve breakthroughs comparable to those it has enabled in text and image generation.

\end{document}